\documentclass[nohyperref]{article}

\usepackage{microtype}
\usepackage{graphicx}
\usepackage{subfigure}
\usepackage{booktabs} 

\usepackage{hyperref}

\usepackage[accepted]{icml2022}

\usepackage{amsmath}
\usepackage{amssymb}
\usepackage{mathtools}
\usepackage{amsthm}

\usepackage[capitalize,noabbrev]{cleveref}

\usepackage{xcolor}
\usepackage{tabularx}
\usepackage{multirow}

\usepackage{amsfonts}
\usepackage{pgfplots}
\usepackage{pgfplotstable}
\usepackage{pifont}

\usepackage{colortbl}
\crefname{section}{§}{§§}
\Crefname{section}{§}{§§}

\theoremstyle{plain}

\theoremstyle{definition}

\theoremstyle{remark}

\usepackage[textsize=tiny]{todonotes}

\icmltitlerunning{Black-Box Tuning for Language-Model-as-a-Service}

\begin{document}

\twocolumn[
\icmltitle{Black-Box Tuning for Language-Model-as-a-Service}



\icmlsetsymbol{equal}{*}

\begin{icmlauthorlist}
\icmlauthor{Tianxiang Sun}{fudan}
\icmlauthor{Yunfan Shao}{fudan}
\icmlauthor{Hong Qian}{ecnu}
\icmlauthor{Xuanjing Huang}{fudan}
\icmlauthor{Xipeng Qiu}{fudan,pengcheng}
\end{icmlauthorlist}

\icmlaffiliation{fudan}{Fudan University}
\icmlaffiliation{pengcheng}{Peng Cheng Laboratory}
\icmlaffiliation{ecnu}{East China Normal University}

\icmlcorrespondingauthor{Tianxiang Sun}{txsun19@fudan.edu.cn}
\icmlcorrespondingauthor{Xipeng Qiu}{xpqiu@fudan.edu.cn}

\icmlkeywords{Language Models, Derivative-Free Optimization}

\vskip 0.3in
]



\printAffiliationsAndNotice{}  

\begin{abstract}
Extremely large pre-trained language models (PTMs) such as GPT-3 are usually released as a service. It allows users to design task-specific prompts to query the PTMs through some black-box APIs. In such a scenario, which we call Language-Model-as-a-Service (LMaaS), the gradients of PTMs are usually unavailable. Can we optimize the task prompts by only accessing the model inference APIs? This paper proposes the \emph{black-box tuning} framework to optimize the continuous prompt prepended to the input text via derivative-free optimization. Instead of optimizing in the original high-dimensional prompt space, which is intractable for traditional derivative-free optimization, we perform optimization in a randomly generated subspace due to the low intrinsic dimensionality of large PTMs. The experimental results show that the black-box tuning with RoBERTa on a few labeled samples not only significantly outperforms manual prompt and GPT-3's in-context learning, but also surpasses the gradient-based counterparts, i.e., prompt tuning and full model tuning.
\end{abstract}

\section{Introduction}
Scaling pre-trained language models (PTMs) has shown increasing power on a wide range of NLP tasks~\cite{Devlin2019BERT,Raffel2020T5,Brown2020GPT3,Fedus2021Switch,Zhang2020CPM,Zhang2021CPM2,Zeng2021PanGu,Sun2021ERNIE3,Qiu2020survey}. Extremely large PTMs can easily generalize to various downstream tasks with a few labeled samples~\cite{Brown2020GPT3}. However, making these large PTMs benefit everyone is a challenge. On the one hand, running such models can be very expensive or even infeasible for most users. On the other hand, the model parameters are often not open-sourced due to commercial considerations and the potential risk of misuse.\footnote{\url{https://openai.com/blog/openai-api/}} Therefore, large PTMs such as GPT-3~\cite{Brown2020GPT3}, ERNIE 3.0~\cite{Sun2021ERNIE3} and Yuan 1.0~\cite{Wu2021Yuan} are usually released as a service, allowing users to access these powerful models through black-box APIs. 

\begin{figure}[t!]
    \centering
    \includegraphics[width=\linewidth]{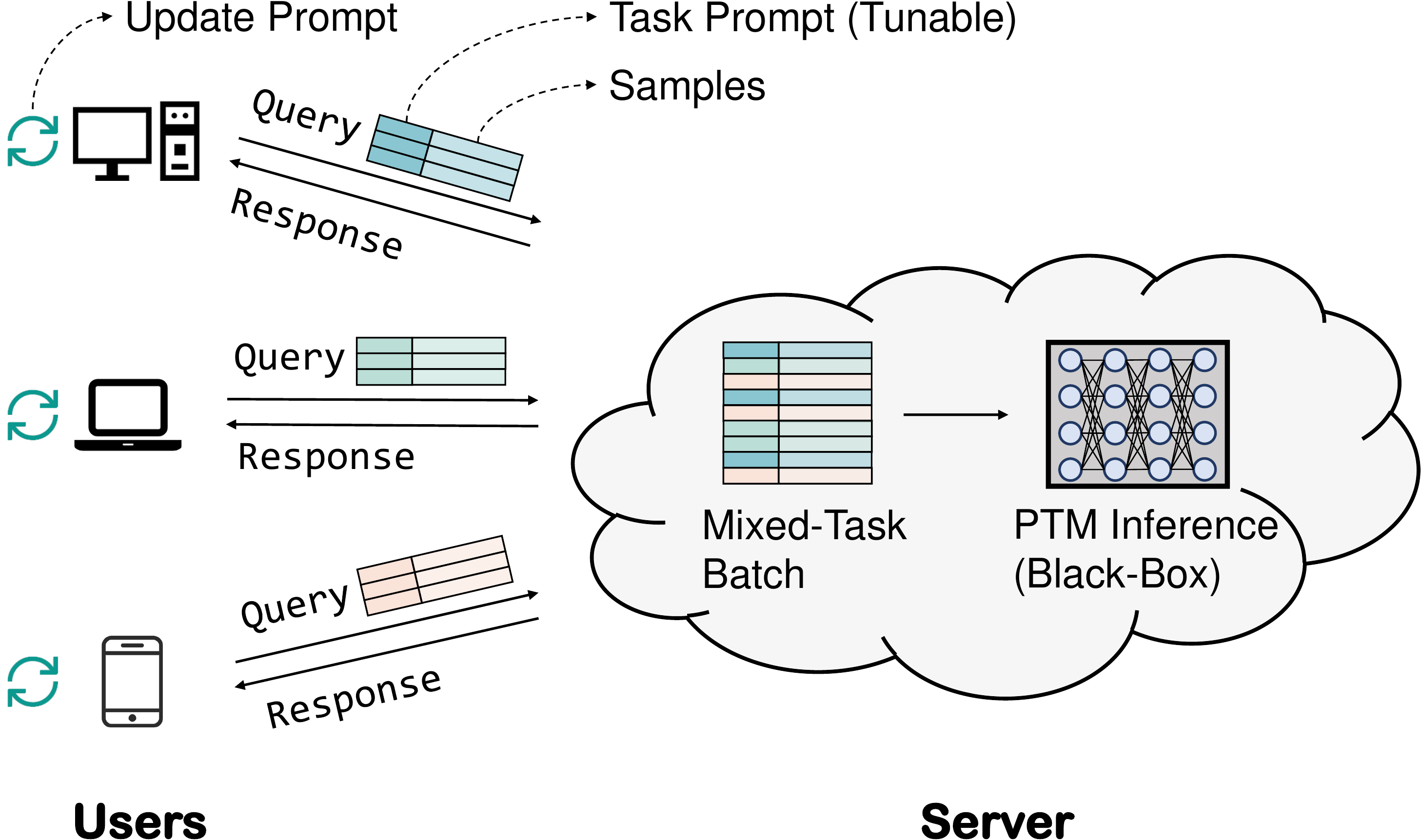}
    \caption{Illustration of Language-Model-as-a-Service (LMaaS). Users can query the PTM deployed on the server through a black-box API. In each query, users can input a task prompt and a batch of texts. On the server side, the samples can be mixed in a large batch to be fed into the PTM. By iteratively querying the PTM through the black-box API, users can optimize and finally obtain good prompts to solve the language tasks of interest.}
    \label{fig:illus}
\end{figure}

In this scenario, called Language-Model-as-a-Service (LMaaS), users can solve the language tasks of interest using the black-box APIs by crafting task-specific text prompts or including training samples in the input texts (a.k.a. in-context learning~\cite{Brown2020GPT3}). Due to the great power of the general-purpose PTMs underlying the APIs, such approaches can achieve considerable performance on simple language tasks, and therefore have powered many interesting applications\footnote{See \url{https://gpt3demo.com/} for examples.}. However, querying large PTMs through hand-crafted text prompts cannot fully exploit labeled data, resulting in unsatisfactory performance in many use cases. 

Instead of designing discrete text prompts, recently much effort has been devoted to continuous prompt tuning~\cite{Li2021Prefix,Hambardzumyan20WARP,Liu2021PTuning}, which is to optimize the continuous prompt injected to the text while keeping the PTM parameters frozen. Such methods only require storing a small continuous prompt for each task, and therefore are highly deployment-efficient. Besides, tuning the continuous prompt can be as effective as fine-tuning the entire model when the PTM becomes large~\cite{Lester2021Prompt}. However, in all the previous methods, the continuous prompts are learned through backpropagation, which is unavailable in the scenario of LMaaS.

\textbf{Can we optimize the task-specific continuous prompts when we only have access to the PTM inference API?} Since gradients are unavailable, we can only invoke derivative-free optimization (DFO)\footnote{Also termed as black-box, zeroth-order or gradient-free optimization.}~\cite{DFO-siamrev,DFO-book,DFO-review}. 
DFO involves a kind of optimization algorithms that do not depend on gradients, but only relies on function values (or fitness values) of sampled solutions. However, DFO algorithms are known to suffer from slow convergence rate when the dimensionality of the search space is high. Thus, it is intractable to optimize even only the continuous prompts, which can be tens of thousands of parameters, using DFO algorithms.



Fortunately, recent work found that common PTMs, despite their large numbers of parameters, have a very low intrinsic dimensionality~\cite{Aghajanyan2021Intrinsic,Qin2021Exploring}. That means, there exists a low-dimensional reparameterization that is as effective for fine-tuning as the full parameter space. It has been demonstrated that optimizing only hundreds~\cite{Aghajanyan2021Intrinsic} or even dozens~\cite{Qin2021Exploring} of parameters can achieve non-trivial performance. Given that the intrinsic dimensionality of the objective function (in our case is the forward computation of PTMs) is low, the optimization can be effectively solved via DFO algorithms with random embedding~\cite{Wang2016BORE,Qian2016DFOHigh,LethamCRB20}.

Based on the these insights, this paper proposes the \textbf{Black-Box Tuning (BBT)} to solve various language understanding tasks by only accessing the PTM inference API. In particular, we manage to optimize the continuous prompt prepended to the input text by iteratively querying the PTM inference API, as briefly depicted in Figure~\ref{fig:illus}. To handle the high dimensionality of the continuous prompt, we project the original prompt space using a random linear projection onto a much smaller subspace and solve this optimization problem with some derivative-free optimizer in that smaller subsapce. In contrast to conventional fine-tuning methods that can only be performed by the service side, black-box tuning allows users to optimize their task-specific prompts locally on resource-limited devices (even without GPUs). Our experimental results demonstrate that prompting RoBERTa\textsubscript{LARGE}~\cite{Liu2019roberta} using BBT on a few labeled samples not only outperforms manual prompt and in-context learning~\cite{Brown2020GPT3}, but also outperforms its gradient-based counterparts, namely prompt tuning~\cite{Lester2021Prompt} and full model tuning.

The contribution of this paper is three folds:\footnote{Our code is publicly available at \url{https://github.com/txsun1997/Black-Box-Tuning}}
\begin{itemize}
    \item This paper proposes a novel scenario (LMaaS) where one should learn to prompt the PTMs by only accessing their inference APIs.
    \item This paper offers a solution (BBT) for such a scenario to accomplish common language understanding tasks without access to model parameters and gradients, such that large-scale PTMs can better benefit users.
    \item Empirical results show that DFO can successfully deal with real-world language tasks by learning to prompt large-scale PTMs with more than millions of parameters. Thus, this work pioneers the work of optimizing large-scale PTMs through DFO methods.
\end{itemize}

\section{Background}
\paragraph{Large-Scale PTMs as APIs.}
It is a promising way to deploy large-scale PTMs to serve downstream applications by providing general-purpose APIs. \textbf{For the service side}, wrapping the computation of the PTM into an easy-to-use API has become a common practice~\cite{Brown2020GPT3,Sun2021ERNIE3,Wu2021Yuan}. In contrast to training, the inference speed of large-scale PTMs can be highly optimized with acceleration techniques such as ORT and TensorRT. In addition, large-scale PTMs are often not open-sourced due to the commercial reasons and the potential risk of misuse. \textbf{For the user side}, even if the large-scale PTMs are available, it is expensive or even infeasible to locally run them.
Thus, how to exploit the PTM inference API to solve conventional language tasks is a promising direction.


\paragraph{Intrinsic Dimensionality of PTMs.}
The intrinsic dimensionality of an objective function is the minimum number of parameters needed to obtain satisfactory solutions~\cite{Li2018Intrinsic}. In particular, the intrinsic dimensionality indicates the lowest dimensional reparameterization that is as effective for optimizing as the full parameter space. \citet{Li2018Intrinsic} propose to measure the intrinsic dimensionality of neural networks by finding the minimal dimensionality of the subspace that is randomly projected from the full trainable parameters, in which they can optimize the neural networks to achieve satisfactory solutions. \citet{Aghajanyan2021Intrinsic} empirically show that large-scale pre-training implicitly compresses the intrinsic dimensionality of downstream NLP tasks. By tuning only hundreds of parameters that are then randomly projected onto the full parameter space of RoBERTa, they can achieve 90\% performance relative to full model tuning. \citet{Qin2021Exploring} show that intrinsic subspace on various tasks can be compressed to less than 100 dimensions with multi-task supervision. This line of research, along with the work of parameter-efficient tuning~\cite{Houlsby2019Adapter,Li2021Prefix,Lester2021Prompt,Sun2021Paradigm,Hu2021LoRA,He2021Unified}, demonstrate that PTMs can well adapt to downstream tasks by tuning a very small proportion of parameters, which implies the possibility of optimizing large-scale PTMs with derivative-free algorithms.

\paragraph{Prompt-Based Learning.}
Prompt-based learning is to formulate downstream tasks as a (masked) language modeling task, and therefore reduces the gap between PTM pre-training and fine-tuning~\cite{Brown2020GPT3,Schick21PET,Schick21Size,Gao20Making,Sun2021Paradigm}. For instance, one can use BERT~\cite{Devlin2019BERT} to predict whether the sentence "\textit{This is a fantastic movie}" is positive or negative by appending the prompt "\textit{It was} \texttt{[MASK]}" and see if BERT predicts "\textit{great}" or "\textit{terrible}" at the masked position. Note that the prompt is not necessarily discrete, it can also be optimized efficiently in continuous space with gradient descent~\cite{Li2021Prefix,Hambardzumyan20WARP,Qin21Learning,Liu2021PTuning,Zhong21OptiPrompt}. In the case of only tuning the continuous prompt while keeping the parameters of large PTMs untouched, one can retain the efficient serving benefits while matching the performance of full model tuning~\cite{Lester2021Prompt}. Our work also proposes to optimize the continuous prompt while keeping the PTM parameters unchanged, but without gradient descent.

\paragraph{Derivative-Free Optimization.}
Derivative-free optimization (DFO) realizes optimization only via the function values $f(\mathbf{x})$ on the sampled solutions $\mathbf{x}$. Most DFO algorithms share a common structure of sampling-and-updating to enhance the quality of solutions.
Representative DFO algorithms include evolutionary algorithms~\cite{Hansen2003Reducing}, Bayesian optimization~\cite{reviewBO16}, etc.
Due to their ability of addressing complex optimization tasks, DFO algorithms have achieved many impressive applications in automatic machine learning~\cite{SnoekLA-nips12}, reinforcement learning~\cite{salimans2017evolution,hu2017sequential-aaai17}, objective detection~\cite{cvpr-ZhangSVPL15}, etc.

\section{Approach}

\subsection{Problem Formulation}
Common language understanding tasks can be formulated as a classification task, which is to predict for a batch of input texts $X$ the labels $Y$. To solve the target language understanding task with a general-purpose PTM, we should modify $X$ with some template (e.g., adding some trigger words and a special token \texttt{[MASK]} for BERT-like PTMs) and map the labels $Y$ to some words in the PTM vocabulary (e.g., the sentiment label "positive" can be mapped to "great"). The modified inputs and labels are denoted as $\Tilde{X}$ and $\Tilde{Y}$. Assume the BERT-like PTM inference API $f$ takes a continuous prompt $\mathbf{p}$ and a batch of modified texts $\Tilde{X}$ as input, and outputs the logits on the masked positions, i.e., $\mathbf{\hat{Y}}=f(\mathbf{p};\Tilde{X})$. With the output logits, we can calculate the loss on this batch of data, which is not necessarily to be differentiable. Our goal is to find the optimal prompt $\mathbf{p}^\star = \arg\min_{\mathbf{p}\in\mathcal{P}}\mathcal{L}(f(\mathbf{p};\Tilde{X}), \Tilde{Y})$, where $\mathcal{P}$ is some search space of interest and $\mathcal{L}$ is some loss function such as negative accuracy. The black-box function $f$ is not available to the optimizer in closed form, but can be evaluated at a query point $(\mathbf{p}; \Tilde{X})$.

\begin{figure*}[t!]
    \centering
    \includegraphics[width=.8\linewidth]{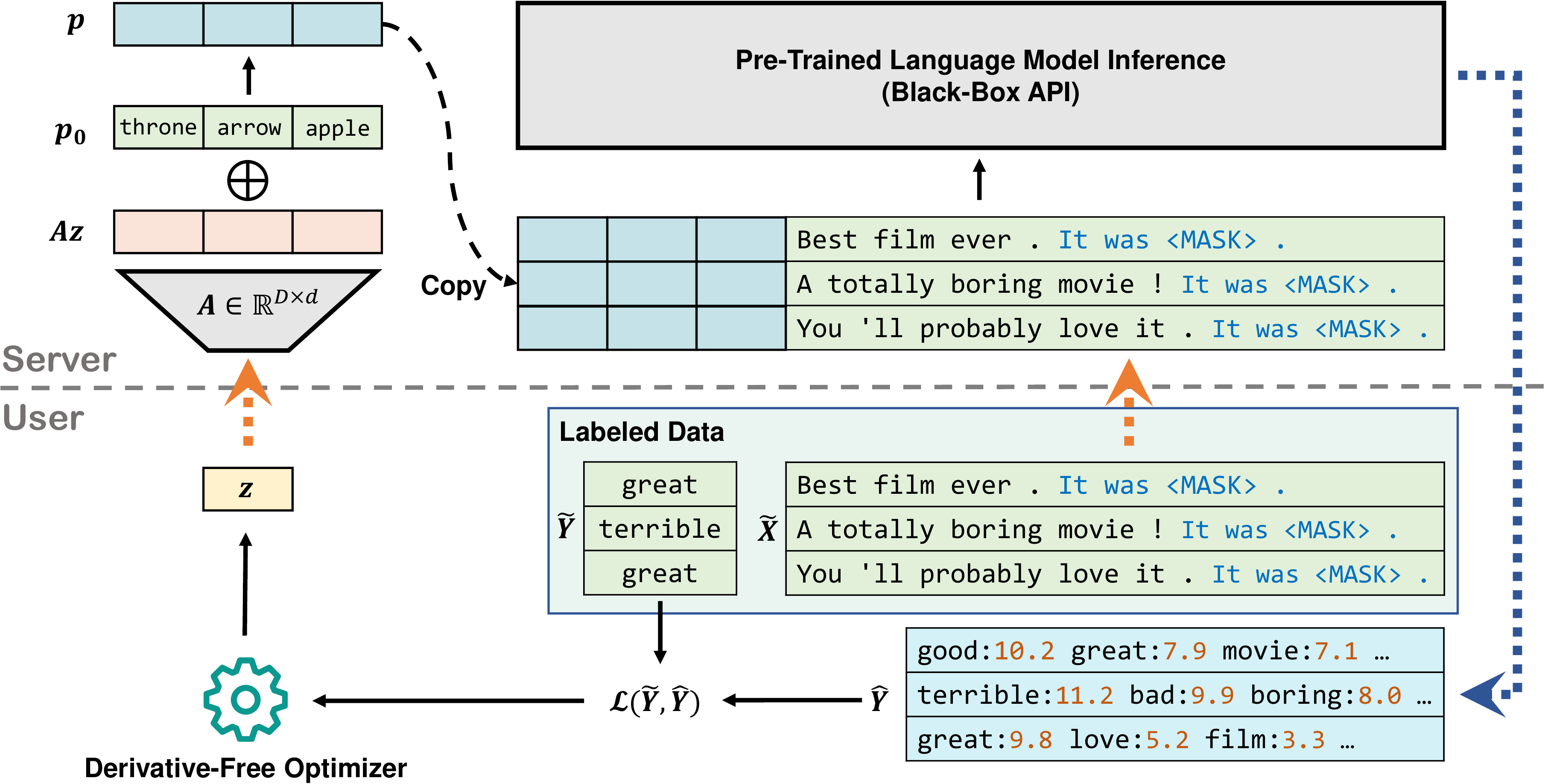}
    \caption{A single iteration of the optimization. Given $\mathbf{z}\in\mathbb{R}^d$ provided by the derivative-free optimizer, we project it to the prompt space by a random matrix $\mathbf{A}\in\mathbb{R}^{D\times d}$. By adding the projected prompt embeddings $\mathbf{Az}$ with some initial prompt embeddings $\mathbf{p}_0$ (in this illustration are the embeddings of tokens randomly sampled from the PTM's vocabulary), we obtain the final prompt embeddings that are then concatenated with the input texts $\Tilde{X}$. By calling the black-box API $f$, which implements the forward computation of the PTM, the predictions on the masked positions are obtained, i.e., $\mathbf{\hat{Y}}=f(\mathbf{p};\Tilde{X})$. With the prediction $\mathbf{\hat{Y}}$ and the golden labels $\Tilde{Y}$ at hand, we can calculate the loss that is used by the derivative-free optimizer to suggest a new $\mathbf{z}$.}
    \label{fig:method}
\end{figure*}

\subsection{Black-Box Tuning}
As demonstrated by~\citet{Lester2021Prompt}, dozens of prompt tokens are required to obtain a competitive performance when only tuning continuous prompts. Given that the embedding dimensionality of large-scale PTMs is usually larger than one thousand (e.g., the word embeddings of RoBERTa\textsubscript{LARGE} are 1024-dimensional), the dimensionality of the continuous prompt $\mathbf{p}\in\mathbb{R}^D$ that we are interested to optimize can be tens of thousands, which makes derivative-free optimization intractable. 
To handle this high-dimensional optimization, since large-scale PTMs have a low intrinsic dimensionality~\cite{Aghajanyan2021Intrinsic,Qin2021Exploring}, we manage to optimize $\mathbf{z}\in\mathbb{R}^d$ in a much smaller subspace ($d\ll D$), and use a random projection matrix $\mathbf{A}\in\mathbb{R}^{D\times d}$ to project $\mathbf{z}$ on the original prompt space $\mathcal{P}$. 


Note that directly projecting $\mathbf{z}$ onto the prompt space that is compatible with the PTM is non-trivial. To ease the optimization, we instead optimize the increment of some initial prompt $\mathbf{p}_0$. For simplicity, we randomly sample $n$ tokens from the PTM vocabulary as initialization. Thus, our objective becomes
\begin{align}
    \mathbf{z}^\star = \mathop{\arg\min}_{\mathbf{z}\in \mathcal{Z}}\mathcal{L}(f(\mathbf{Az}+\mathbf{p}_0;\Tilde{X}), \Tilde{Y})\,,
\end{align}
where $\mathcal{Z}$ is the search space. Previous work~\cite{Wang2016BORE,Qian2016DFOHigh,LethamCRB20} in derivative-free optimization usually sets each entry in the random matrix $\mathbf{A}$ by sampling from some normal distribution. However, this sampling strategy does not perform well in our scenario. Instead, we set values of the random matrix $\mathbf{A}$ by sampling from a uniform distribution adopted in~\citet{He2015Delving} (cf. Appendix~\ref{append:add_exp} for the comparison). We restrict the search space to $\mathcal{Z}=[-5, 5]^d$.

For the loss function $\mathcal{L}$, a straightforward alternative is using negative accuracy. However, the reward of accuracy can be sparse and less informative, especially when training data is limited. Thus, we also consider two loss functions that are more sensitive to predictions, i.e., cross entropy and hinge loss. Given the output logits $\mathbf{\hat{y}}$ over a candidate set of label words, and the golden label word $\Tilde{y}$ of a certain sample, the cross entropy is defined as
\begin{align}
    \mathcal{L}_{\text{CE}}(\mathbf{\hat{y}}, \Tilde{y}) = -\log \text{Softmax}_{\Tilde{y}}(\mathbf{\hat{y}}).
\end{align}
For hinge loss, we adopt a multi-class extension~\cite{Weston1999SVMMC},
\begin{align}
    \mathcal{L}_{\text{Hinge}}(\mathbf{\hat{y}}, \Tilde{y}) = \sum_{i\neq \Tilde{y}}\max (0, \gamma + \mathbf{\hat{y}}_i - \mathbf{\hat{y}}_{\Tilde{y}}).
\end{align}
In this work we set the margin $\gamma=2$. The performances of using cross entropy, hinge loss, and negative accuracy are compared in Figure~\ref{fig:ablation}.

\begin{table*}[t!]
\centering
\caption{Statistics, manual templates, and label words used in our experiments. $\mid\mathcal{Y}\mid$: number of classes.}
\label{tab:data}
\resizebox{\linewidth}{!}{
\begin{tabular}{llcrrcll}
\toprule
\textbf{Category} & \textbf{Dataset} & \textbf{$\mid\mathcal{Y}\mid$} & \textbf{$\mid$Train$\mid$} & \textbf{$\mid$Test$\mid$} & \textbf{Type} & \textbf{Template} & \textbf{Label words} \\ \midrule
\multirow{6}{*}{\begin{tabular}[c]{@{}l@{}}single-\\ sentence\end{tabular}}
& SST-2 & 2 & 67k & 0.9k & sentiment & $\langle S\rangle$. It was \texttt{[MASK]}. & great, bad \\
& Yelp P. & 2 & 560k & 38k & sentiment & $\langle S\rangle$. It was \texttt{[MASK]}. & great, bad \\
& AG's News & 4 & 120k & 7.6k & topic & \texttt{[MASK]} News: $\langle S\rangle$ & World, Sports, Business, Tech \\
& DBPedia & 14 & 560k & 70k & topic & [Category: \texttt{[MASK]}] $\langle S\rangle$ & Company, Education, Artist, Athlete, Office,\\ 
& & & & & & & Transportation, Building, Natural, Village,\\
& & & & & & & Animal, Plant, Album, Film, Written\\ \midrule
\multirow{3}{*}{\begin{tabular}[c]{@{}l@{}}sentence-\\ pair\end{tabular}}
& MRPC & 2 & 3.7k & 0.4k & paraphrase & $\langle S_1\rangle$ ? \texttt{[MASK]}, $\langle S_2\rangle$ & Yes, No \\
& RTE & 2 & 2.5k & 0.3k & NLI & $\langle S_1\rangle$ ? \texttt{[MASK]}, $\langle S_2\rangle$ & Yes, No \\
& SNLI & 3 & 549k & 9.8k & NLI & $\langle S_1\rangle$ ? \texttt{[MASK]}, $\langle S_2\rangle$ & Yes, Maybe, No \\ \bottomrule
\end{tabular}
}
\end{table*}

\subsection{The CMA Evolution Strategy}
As demonstrated in~\citet{Aghajanyan2021Intrinsic}, the intrinsic dimensionality of PTMs like RoBERTa\textsubscript{LARGE} on various tasks can be hundreds. To handle optimization of such scale, we adopt the CMA-ES (Covariance Matrix Adaptation Evolution Strategy)~\cite{Hansen2001CMA,Hansen2003Reducing}, which is a widely used evolutionary algorithm for non-convex black-box optimization in continuous domain.



In particular, CMA-ES maintains a parameterized search distribution model, i.e., multivariate normal distribution. In each iteration, CMA-ES samples a population of new query solutions (also referred to as individuals or offspring) from the multivariate normal distribution model
\begin{align}
    \mathbf{z}_i^{(t+1)} \sim \mathbf{m}^{(t)} + \mathbf{\sigma}^{(t)}\mathcal{N}(\mathbf{0}, \mathbf{C}^{(t)})\,,
\end{align}
where $i=1,\dots,\lambda$ and $\lambda$ is the population size. $\mathbf{m}^{(t)}\in \mathbb{R}^d$ is the mean vector of the search distribution at iteration step $t$, $\sigma ^{(t)}\in\mathbb{R}_+$ is the overall standard deviation that controls the step length, and $\mathbf{C}^{(t)}\in \mathbb{R}^{d\times d}$ is the covariance matrix that determines the shape of the distribution ellipsoid. By maximizing the likelihood of successful steps, $\mathbf{m}^{(t)}$, $\sigma^{(t)}$, $\mathbf{C}^{(t)}$ are updated (cf.~\citet{Hansen16a2016CMATutorial} for more details). 

\subsection{Pre-Training Prompt Embedding}
\label{sec:ppe}

Considering that sentence-pair tasks can share the same template and label words, as shown in Table~\ref{tab:data}, we can pre-train a prompt embedding $\mathbf{p}_0$ on some publicly available NLI task (in our experiments we use the MNLI~\cite{Williams2018MNLI} training set) for a better initialization. For other classification tasks we set $\mathbf{p}_0$ as word embeddings randomly drawn from the vocabulary of RoBERTa\textsubscript{LARGE}. 

\section{Experiments}
\subsection{Setup}
\label{sec:setup}
\paragraph{Dataset.}
We conduct experiments on several common language understanding tasks including sentiment analysis, topic classification, natural language inference (NLI), and paraphrase. For sentiment analysis, we choose SST-2~\cite{Socher2013SST} and Yelp polarity~\cite{Zhang2015Char}. For topic classification, we choose AG's News and DBPedia~\cite{Zhang2015Char}. For NLI, we choose SNLI~\cite{Bowman2015SNLI} and RTE~\cite{Wang2019GLUE}. For paraphrase, we choose MRPC~\cite{Dolan2005MRPC}. The statistics, manual templates and label words of these datasets are shown in Table~\ref{tab:data}.

\paragraph{Few-Shot Setting.}
For a broad range of users, the amount of labeled data can be limited, in which case they can resort to the deployed large PTMs due to their great power of few-shot learning~\cite{Brown2020GPT3}. Hence, in this paper we conduct experiments in the few-shot setting. We randomly select $k$ samples for each class to construct a $k$-shot training set $\mathcal{D}_{\text{train}}$, and compose a development set $\mathcal{D}_{\text{dev}}$ by randomly drawing another $k$ samples from the original training set and ensure that $|\mathcal{D}_{\text{train}}|=|\mathcal{D}_{\text{dev}}|$ to simulate the true few-shot learning setting~\cite{Perez2021TrueFewShot}. 
Following~\citet{Zhang2021Revisiting}, \citet{Gao20Making}, and \citet{Gu2021PPT}, we use the original development sets as the test sets. For datasets without development sets, we use the original test sets. Hence, in our experiments $|\mathcal{D}_{\text{test}}| \gg |\mathcal{D}_{\text{train}}| = |\mathcal{D}_{\text{dev}}|$.

\paragraph{Backbone Model.}
We choose RoBERTa\textsubscript{LARGE}~\cite{Liu2019roberta} as our backbone model because: (1) We mainly focus on language understanding tasks; (2) \citet{Aghajanyan2021Intrinsic} have demonstrated that RoBERTa\textsubscript{LARGE} has a very small intrinsic dimensionality (about hundreds) on many tasks. It is worth noting that generative PTMs such as GPT~\cite{Brown2020GPT3}, T5~\cite{Raffel2020T5} and BART~\cite{Lewis2020BART} are also compatible with our framework if we convert downstream tasks into a unified text-to-text format. We leave for future work the applications of generative PTMs.

\begin{table}[t]
\centering
\caption{Default configuration of hyper-parameters.}
\label{tab:hyperparam}
\resizebox{.7\linewidth}{!}{
\begin{tabular}{lc}
\toprule
\textbf{Hyper-parameter} & \textbf{Default} \\ \midrule
Prompt length ($L$)              & 50         \\
Subspace dimension ($d$)         & 500        \\
Population size ($\lambda$)      & 20         \\
Random projection ($\mathbf{A}$) & Uniform    \\
Loss function $\mathcal{L}$      & Cross Entropy \\ 
Budget (\# of API calls)         & 8000      \\ \bottomrule
\end{tabular}
}
\vskip -0.2in
\end{table}

\paragraph{Baselines.}
We compare our proposed black-box tuning with two kinds of methods: gradient-based methods and gradient-free methods. 
\textit{For gradient-based methods}, we consider three baselines: 
\textbf{(1) Prompt Tuning}: Following \citet{Lester2021Prompt}, we only train the continuous prompts prepended to the input texts while keeping the PTM frozen. We use an Adam optimizer~\cite{kingma2015adam} with learning rate of 5e-4 and batch size of 16 for 1000 epochs. For fair comparison, we use the same prompt length, manual template, label words, and the same pre-trained prompt embedding for initialization on sentence-pair tasks as black-box tuning.
\textbf{(2) P-Tuning v2}~\cite{Liu2021PTuningv2} is an improved variant of prompt tuning. Instead of injecting continuous prompts merely into the input layer, P-Tuning v2 prepends and optimizes continuous prompts at every layer of the PTM. We optimize the prompts of length 128 at each layer using an Adam optimizer with learning rate of 5e-4 and batch size of 32 for 2000 epochs.
\textbf{(3) Model Tuning}: We fine-tune the entire PTM on each task using an Adam optimizer with learning rate of 1e-5 and batch size of 16 for 200 epochs.
\textit{For gradient-free methods}, we consider three baselines: 
\textbf{(1) Manual Prompt}: We directly use the templates and label words in Table~\ref{tab:data} to conduct zero-shot evaluation. The results of manual prompt can be seen as initial points of our method.
\textbf{(2) In-context Learning}: Following \citet{Brown2020GPT3}, we randomly select up to 32 training samples and concatenate them with the input texts.
\textbf{(3) Feature-based Methods}: Feature-based methods~\cite{Peters2019ToTune} is also a competitive baseline for LMaaS, where one can request the features encoded by the large PTM and locally train a classifier to accomplish the task of interest. Here we consider two implementations:
\textbf{(a) Feature-MLP}: We train a two-layered MLP classifier on the \texttt{[CLS]} representation of the PTM.
\textbf{(b) Feature-BiLSTM}: We train a bidirectional LSTM~\cite{Hochreiter1997LSTM} on the representations of the sequence of tokens, followed by a linear classifier on the top. For both implementations of feature-based methods, we use an Adam optimizer with learning rate of 3e-4 and batch size of 16 to train the attached classifiers for 1000 epochs. For black-box tuning, we give in Table~\ref{tab:hyperparam} the default configuration of hyper-parameters used in our experiments. The effect of each hyper-parameter is explored in \cref{sec:ablation}. 

\begin{table*}[t!]
\centering
\caption{Overall comparison on various language understanding tasks. We report mean and standard deviation of performance over 3 different splits (\cref{sec:setup}). All of the results are obtained with pre-trained RoBERTa\textsubscript{LARGE} in 16-shot (per class) setting.}
\label{tab:main_results}
\resizebox{\linewidth}{!}{
\begin{tabular}{lcccccccr}
\toprule
\multirow{2}{*}{\textbf{Method}}                                                  & \textbf{SST-2} & \textbf{Yelp P.} & \textbf{AG's News} & \textbf{DBPedia} & \textbf{MRPC} & \textbf{SNLI} & \textbf{RTE} & \multirow{2}{*}{\textbf{Avg.}} \\
                                                                                  & acc            & acc              & acc                & acc              & F1            & acc           & acc          & \\ \midrule
\multicolumn{9}{c}{\textit{Gradient-Based Methods}}                                                                                                                                                                      \\ \midrule
Prompt Tuning & 68.23 \small{$\pm$3.78} & 61.02 \small{$\pm$6.65} & 84.81 \small{$\pm$0.66} & 87.75 \small{$\pm$1.48} & 51.61 \small{$\pm$8.67} & 36.13 \small{$\pm$1.51} & 54.69 \small{$\pm$3.79} & 63.46 \\
\ + Pre-trained prompt & / & / & / & / & 77.48 \small{$\pm$4.85} & 64.55 \small{$\pm$2.43} & 77.13 \small{$\pm$0.83} & 74.42\\
P-Tuning v2    & 64.33 \small{$\pm$3.05} &92.63 \small{$\pm$1.39} &83.46 \small{$\pm$1.01} &97.05 \small{$\pm$0.41} &68.14 \small{$\pm$3.89} &36.89 \small{$\pm$0.79} &50.78 \small{$\pm$2.28} & 70.47 \\ 
Model Tuning & 85.39 \small{$\pm$2.84} & 91.82 \small{$\pm$0.79} & 86.36 \small{$\pm$1.85} & 97.98 \small{$\pm$0.14} & 77.35 \small{$\pm$5.70} & 54.64 \small{$\pm$5.29} & 58.60 \small{$\pm$6.21} & 78.88\\
\midrule
\multicolumn{9}{c}{\textit{Gradient-Free Methods}}                                                                                                                                                                      \\ \midrule
Manual Prompt            & 79.82 & 89.65 & 76.96 & 41.33 & 67.40 & 31.11 & 51.62 & 62.56 \\
In-Context Learning      & 79.79 \small{$\pm$3.06} & 85.38 \small{$\pm$3.92} & 62.21 \small{$\pm$13.46} & 34.83 \small{$\pm$7.59} & 45.81 \small{$\pm$6.67} & 47.11 \small{$\pm$0.63} & 60.36 \small{$\pm$1.56} & 59.36\\
Feature-MLP   &64.80 \small{$\pm$1.78} &79.20 \small{$\pm$2.26} &70.77 \small{$\pm$0.67} & 87.78 \small{$\pm$0.61}&68.40 \small{$\pm$0.86} &42.01 \small{$\pm$0.33} &53.43 \small{$\pm$1.57} & 66.63\\
Feature-BiLSTM   &65.95 \small{$\pm$0.99} &74.68 \small{$\pm$0.10} &77.28 \small{$\pm$2.83} &90.37 \small{$\pm$3.10} &71.55 \small{$\pm$7.10} &46.02 \small{$\pm$0.38} &52.17 \small{$\pm$0.25} & 68.29\\
\textbf{Black-Box Tuning} & 89.56 \small{$\pm$0.25} & 91.50 \small{$\pm$0.16} & 81.51 \small{$\pm$0.79} & 87.80 \small{$\pm$1.53} & 61.56 \small{$\pm$4.34} &  46.58 \small{$\pm$1.33}  & 52.59 \small{$\pm$2.21} & 73.01\\
\ + Pre-trained prompt & / & / & / & / & 75.51 \small{$\pm$5.54} & 83.83 \small{$\pm$0.21} & 77.62 \small{$\pm$1.30} & \textbf{83.90} \\ 

\bottomrule
\end{tabular}
}
\end{table*}

\begin{table*}[th]
\centering
\caption{Comparison of deployment efficiency, viability of as-a-service, test accuracy, training time, memory footprint, and the amount of data to be uploaded/downloaded. $^\star$ indicates the training time of the implementation with ONNX Runtime. All the compared methods are performed on the same 16-shot splits of SST-2 and AG's News.}
\label{tab:time_memo}
\resizebox{.95\linewidth}{!}{
\begin{tabular}{lcccccccc}
\toprule
& \textbf{Deployment-} & \textbf{As-A-} & \textbf{Test} & \textbf{Training} & \multicolumn{2}{c}{\textbf{Memory Footprint}} & \textbf{Upload} & \textbf{Download} \\
& \textbf{Efficient} & \textbf{Service} & \textbf{Accuracy}                & \textbf{Time}        & User          & Server        & per query        & per query   \\ \midrule
\multicolumn{9}{c}{SST-2 (max sequence length: 47)}                                                                                         \\ \midrule
Prompt Tuning   & $\surd$ & $\times$ & 72.6 & 15.9 mins & - & 5.3 GB & - & - \\
Model Tuning   & $\times$ & $\times$ & 87.8 & 9.8 mins & - & 7.3 GB & - & - \\
Feature-MLP   & $\surd$ & $\surd$ & 63.8 & 7.0 mins & 20 MB & 2.8 GB & 4 KB & 128 KB \\
Feature-BiLSTM   & $\surd$ & $\surd$ & 66.2 & 9.3 mins & 410 MB & 2.8 GB & 4 KB & 6016 KB \\
Black-Box Tuning  & $\surd$ & $\surd$ & 89.4 & 10.1 (6.1$^\star$) mins & 30 MB & 3.0 GB & 6 KB & 0.25 KB \\ \midrule
\multicolumn{9}{c}{AG's News (max sequence length: 107)}\\ \midrule
Prompt Tuning & $\surd$ & $\times$ & 84.0 & 30.2 mins & - & 7.7 GB & - & - \\
Model Tuning & $\times$ & $\times$ & 88.4 & 13.1 mins & - & 7.3 GB & - & - \\
Feature-MLP & $\surd$ & $\surd$ & 71.0 & 13.5 mins & 20 MB & 3.6 GB & 20 KB & 256 KB \\
Feature-BiLSTM & $\surd$ & $\surd$ & 73.1 & 19.7 mins & 500 MB & 3.6 GB & 20 KB & 27392 KB \\
Black-Box Tuning & $\surd$ & $\surd$ & 82.6 & 21.0 (17.7$^\star$) mins & 30 MB & 4.6 GB & 22 KB & 1 KB \\ \bottomrule
\end{tabular}
}
\end{table*}

\subsection{Results}

\paragraph{Overall Comparison.}
We first demonstrate the experimental results of black-box tuning and the baselines across 7 datasets in Table~\ref{tab:main_results}. The proposed black-box tuning significantly outperforms the other four gradient-free methods. We observe that in-context learning performs even worse than manual prompt on some tasks, and suffers from high variance. That means, in-context learning cannot effectively utilize labeled samples included in the context. Feature-based methods perform slightly better than manual prompt and in-context learning. Meanwhile, Feature-BiLSTM outperforms Feature-MLP due to its advantage of using more informative features. Surprisingly, black-box tuning also outperforms its gradient-based counterparts, namely prompt tuning, p-tuning v2, and model tuning, on average performance of the 7 tasks. Note that the only difference between prompt tuning and black-box tuning is whether we use gradient descent (i.e., Adam optimizer) or DFO algorithm (i.e., CMA-ES). Based on the experimental results, we suspect that gradient-based optimization tends to overfit the small training data while DFO tends to find better solutions due to its exploration mechanism. In addition, we find that model tuning performs much better than prompt tuning and black-box tuning when number of classes is large (e.g., DBPedia). On NLI tasks (i.e., SNLI and RTE), when using pre-trained prompt embedding (\cref{sec:ppe}), prompt tuning and black-box tuning significantly outperform model tuning, which also confirms the effectiveness of prompt pre-training~\cite{Gu2021PPT} in the context of black-box tuning.

\begin{figure*}[t!]
    \centering
    \subfigure{
    \includegraphics[width=.23\linewidth]{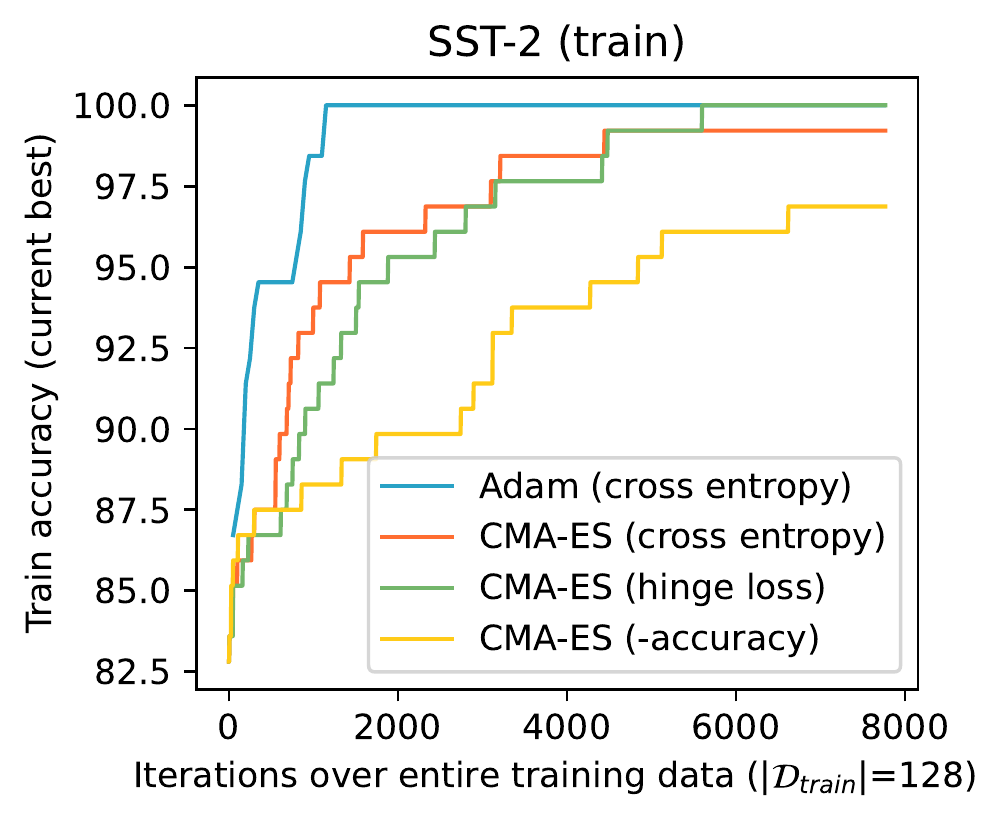}
    }
    \subfigure{
    \includegraphics[width=.23\linewidth]{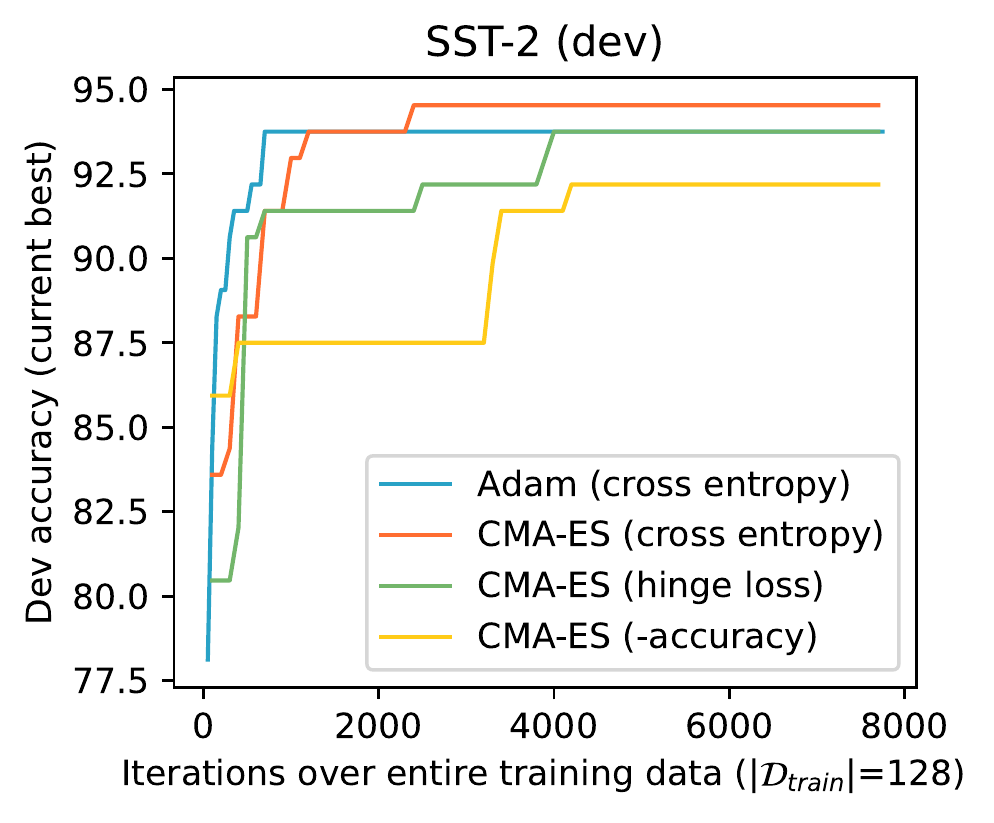}
    }
    \subfigure{
    \includegraphics[width=.23\linewidth]{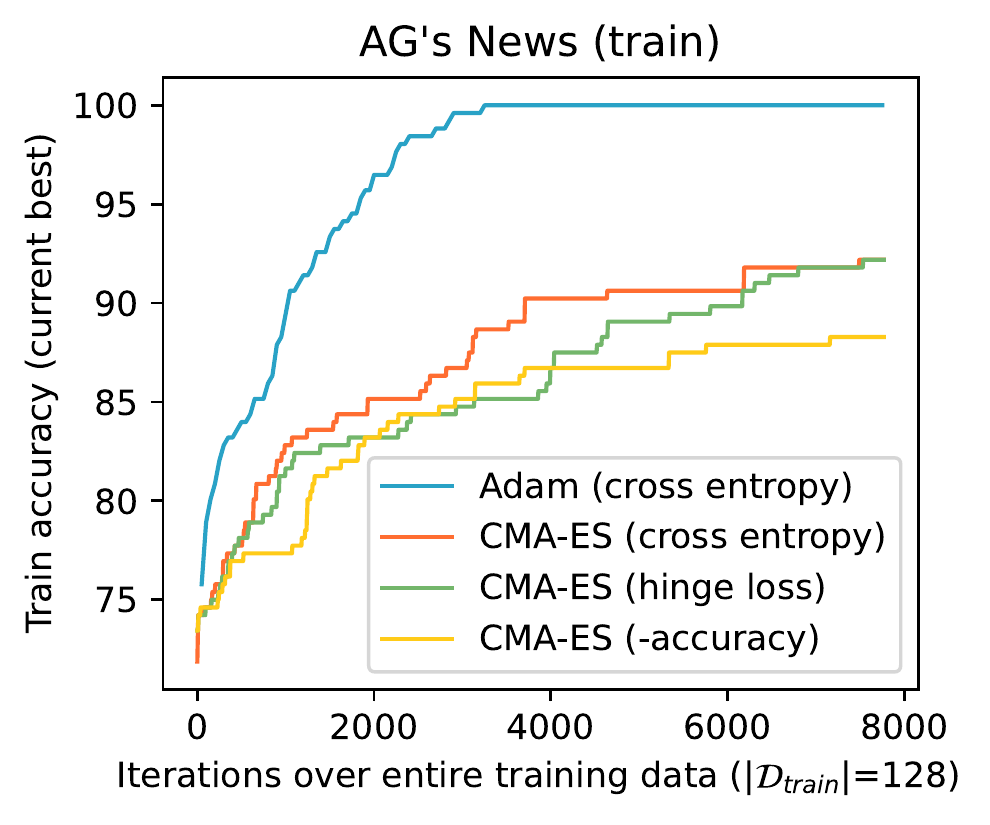}
    }
    \subfigure{
    \includegraphics[width=.23\linewidth]{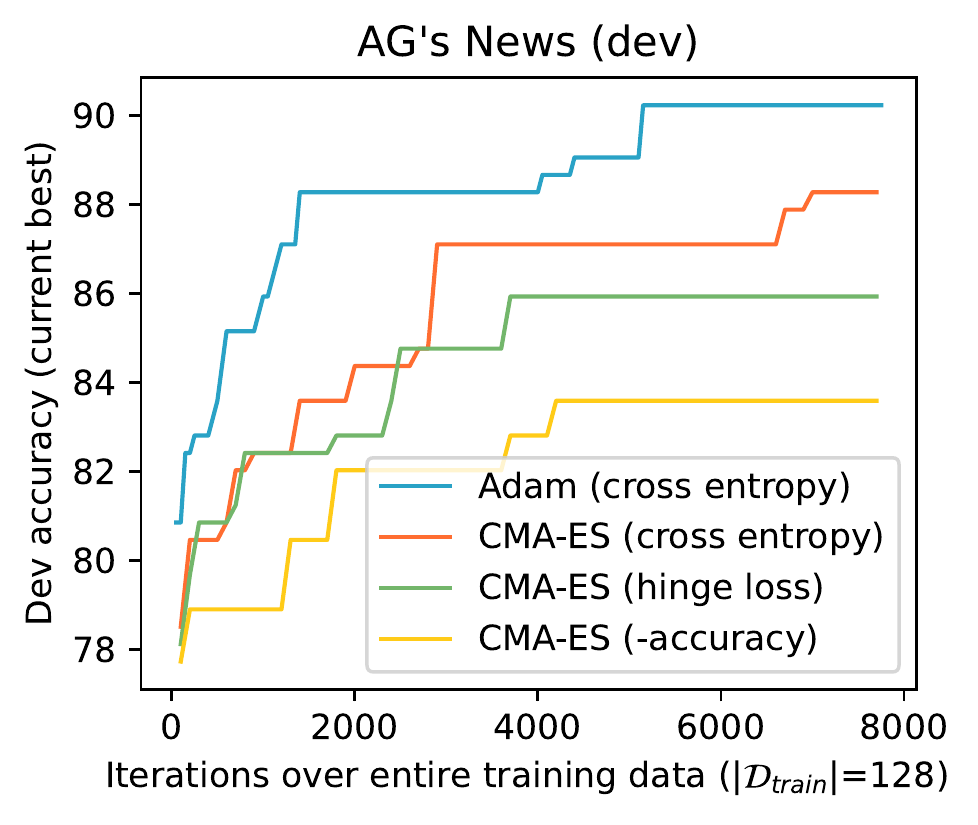}
    }
    \\
    \subfigure{
    \includegraphics[width=.23\linewidth]{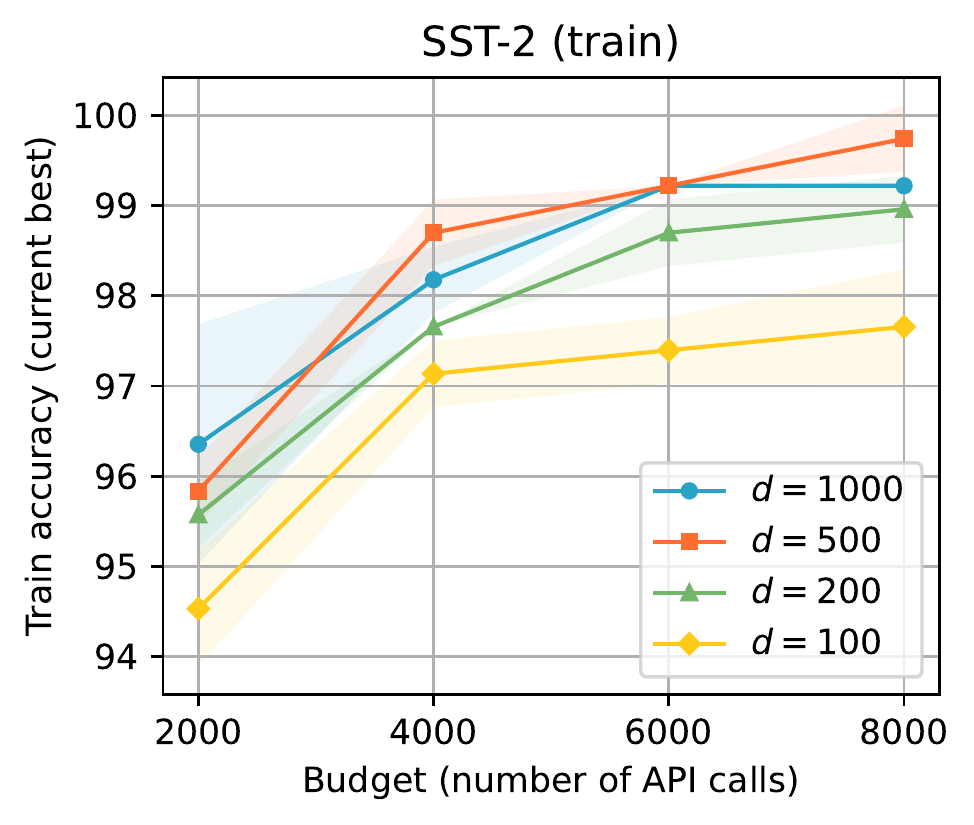}
    }
    \subfigure{
    \includegraphics[width=.23\linewidth]{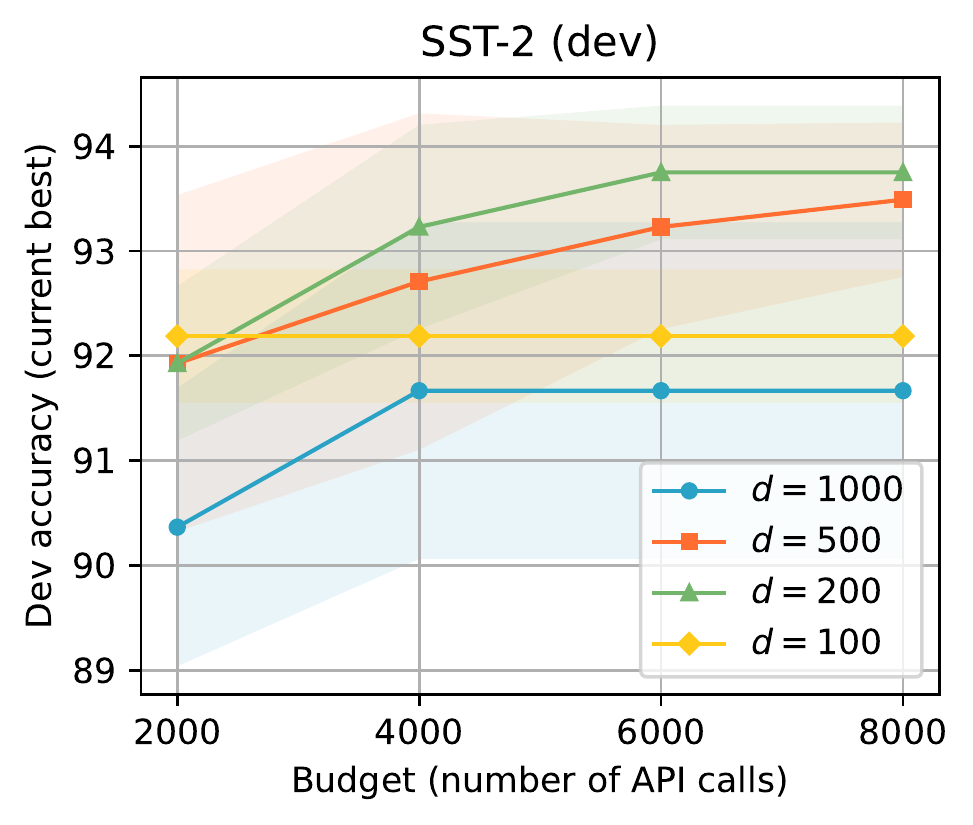}
    }
    \subfigure{
    \includegraphics[width=.23\linewidth]{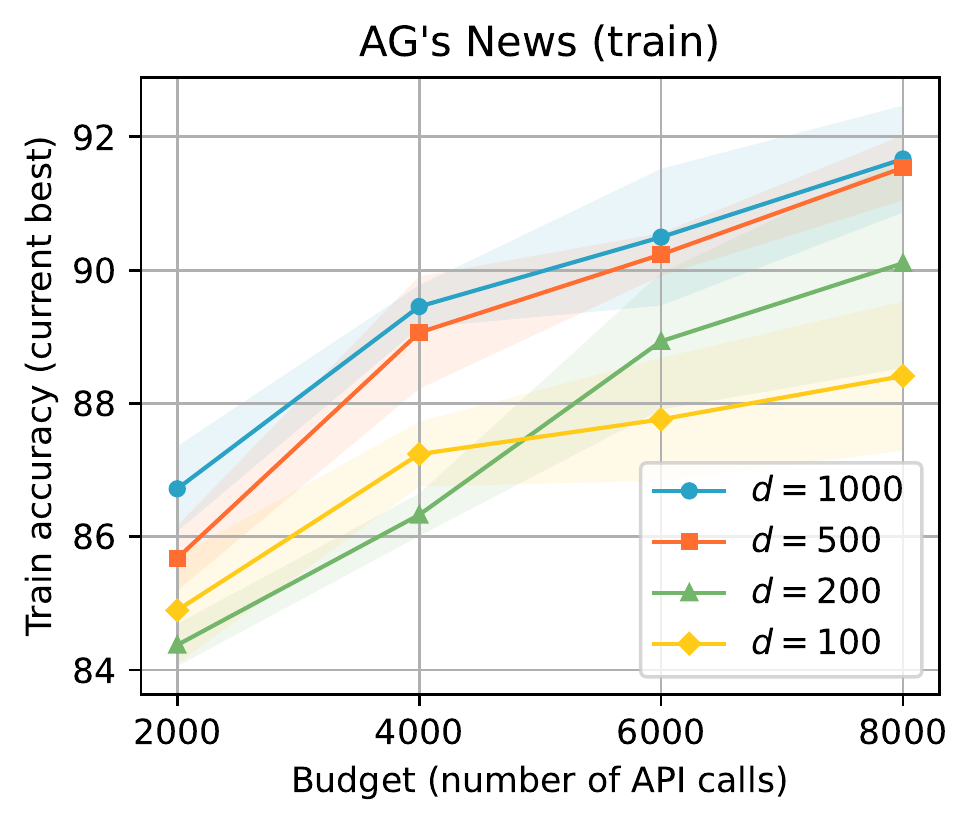}
    }
    \subfigure{
    \includegraphics[width=.23\linewidth]{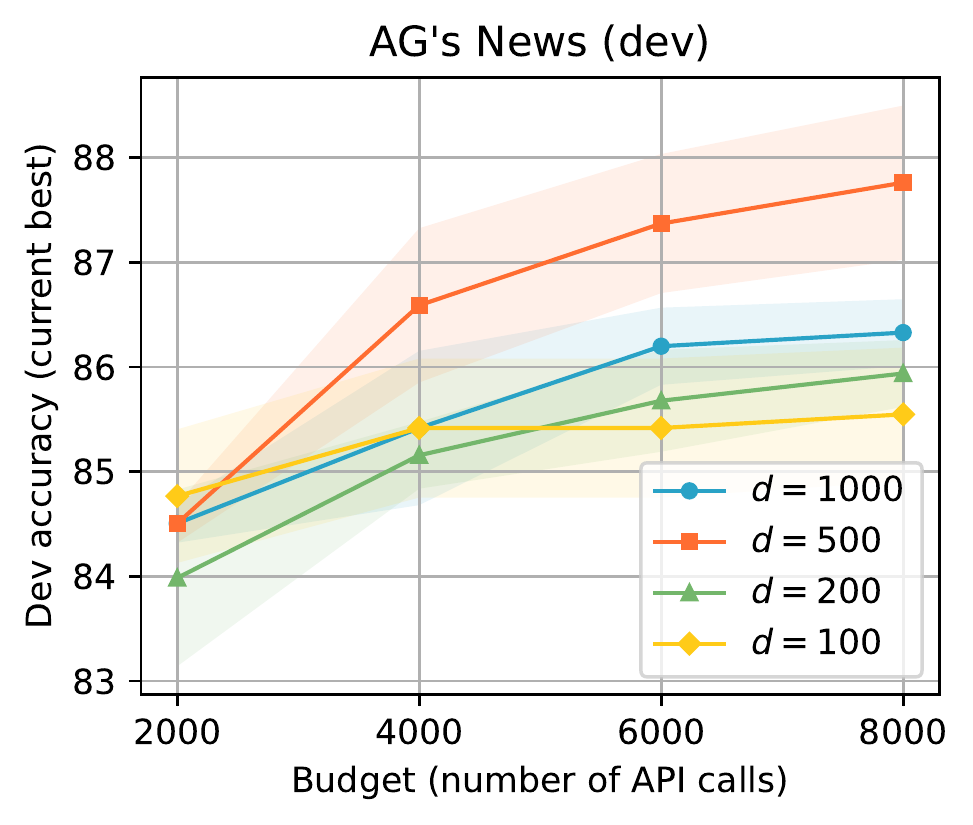}
    }
    \\
    \subfigure{
    \includegraphics[width=.23\linewidth]{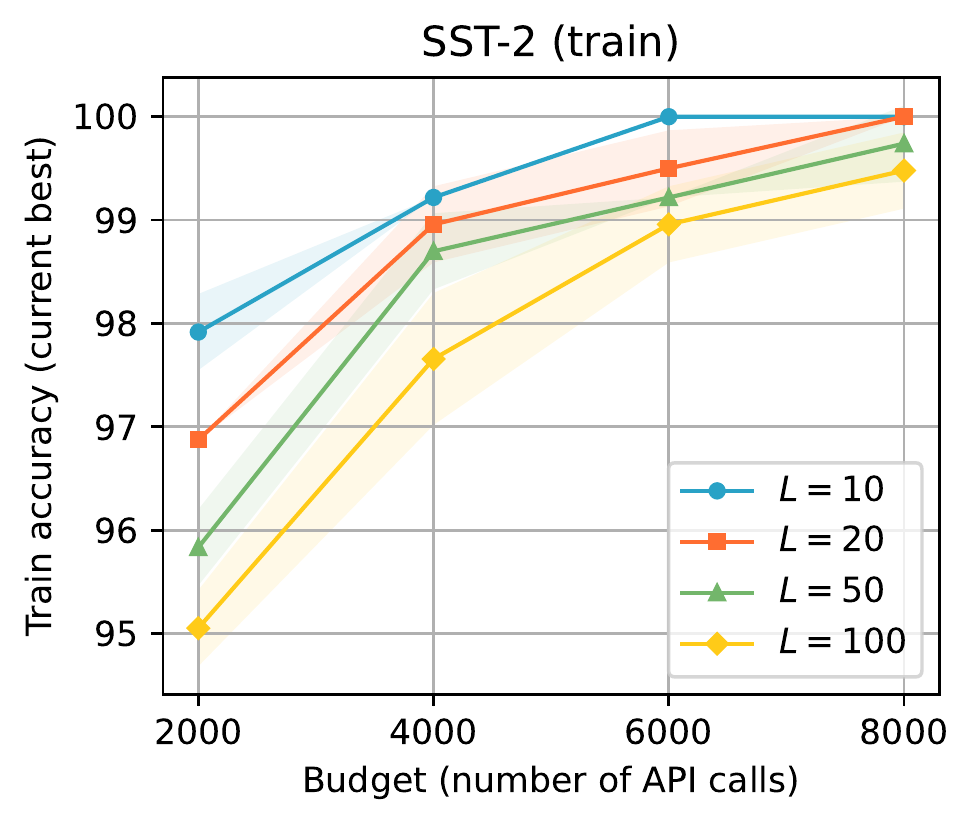}
    }
    \subfigure{
    \includegraphics[width=.23\linewidth]{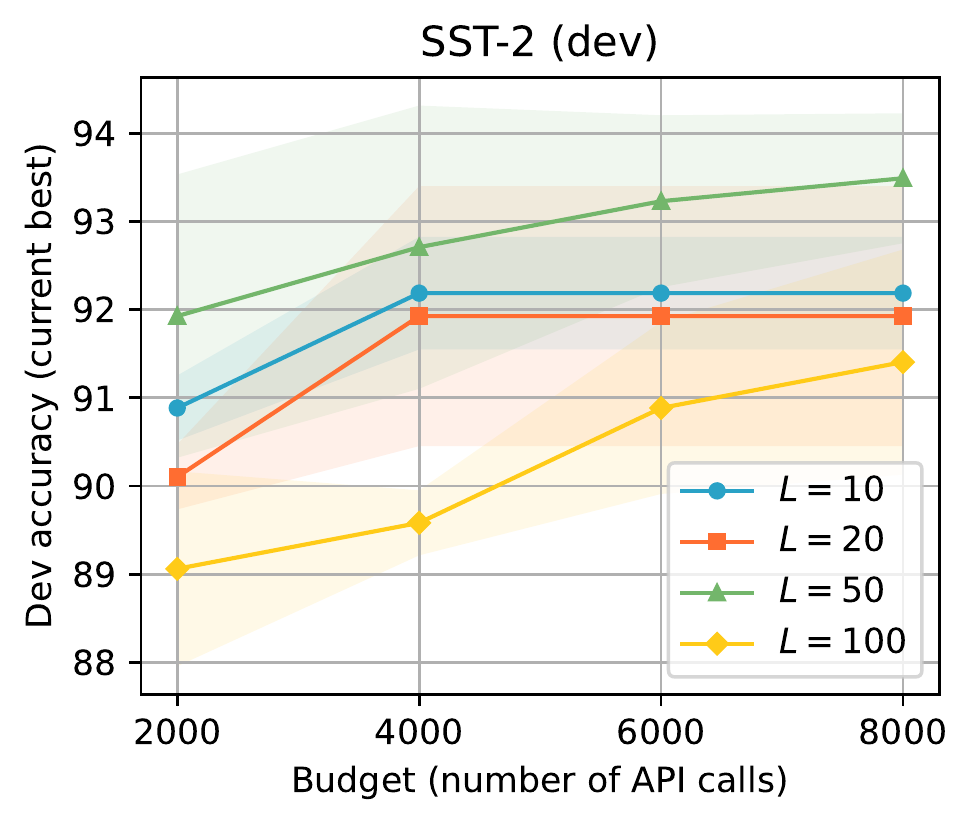}
    }
    \subfigure{
    \includegraphics[width=.23\linewidth]{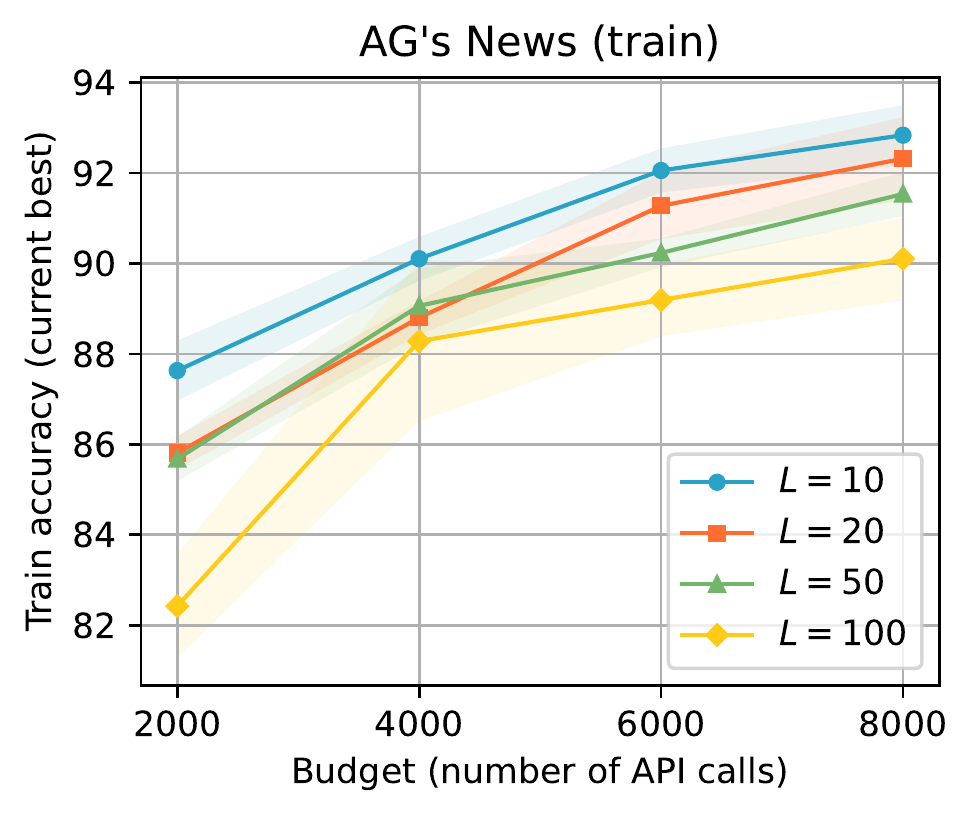}
    }
    \subfigure{
    \includegraphics[width=.23\linewidth]{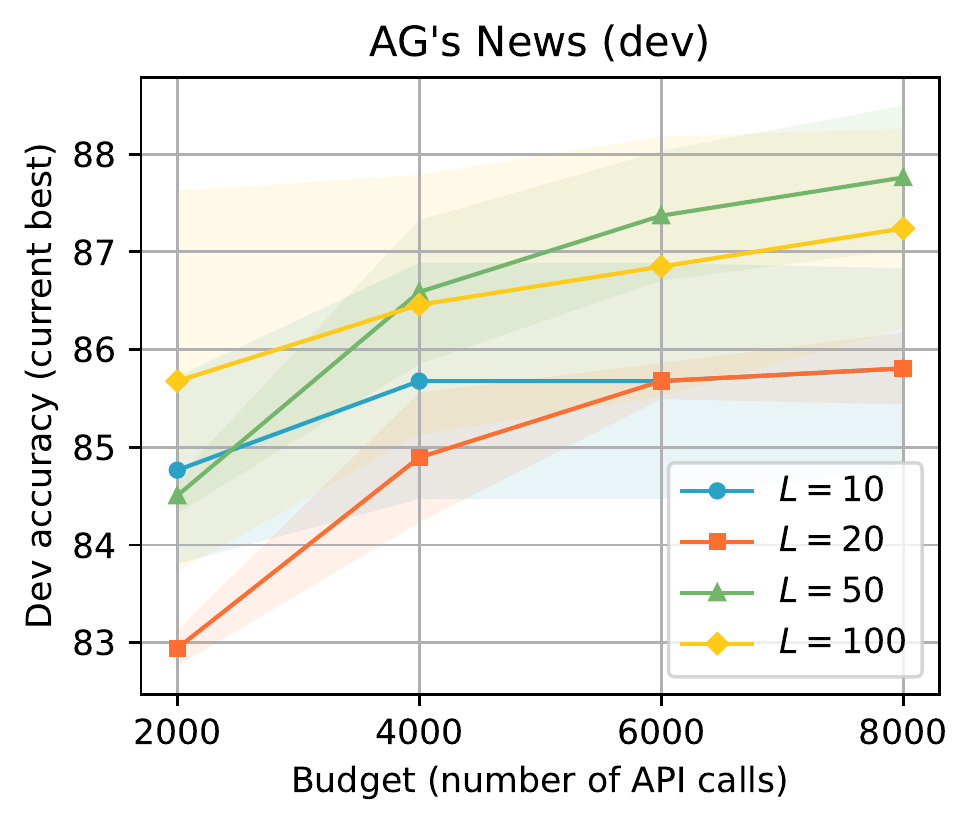}
    }
    \caption{Ablations of loss function, subspace dimensionality, and prompt length. We show mean and standard deviation of performance over 3 runs with different random seeds. Ablations of the random projection and the population size can be found in Appendix~\ref{append:add_exp}.}
    \label{fig:ablation}
\end{figure*}

\paragraph{Detailed Comparison.}
In the scenario of LMaaS, there are many other factors to be considered. In Table~\ref{tab:time_memo} we compare black-box tuning and the baseline methods in terms of deployment efficiency, viability of as-a-service, training time, memory usage on the user side and the server side, and the amount of data to be uploaded and downloaded. Model tuning is not deployment-efficient because it needs to maintain a copy of the entire model for each user. Gradient-based methods cannot make the PTM serve as a service due to the requirement of gradients. Feature-based methods and black-box tuning are suitable for LMaaS. However, feature-based methods cannot achieve competitive results when labeled data is limited. Therefore, among all the considered methods, only black-box tuning can achieve satisfactory performance while maintaining reasonable training time, memory footprint, and network load. Unlike gradient-based methods, in which the optimization cost is proportional to the size of the PTM, the optimization cost of black-box tuning is decoupled from the scale of the PTM, and only relies on the subspace dimensionality. 
For fair comparison of training time, we perform early stopping for all the compared methods, i.e., we stop learning if the development accuracy does not increase after 1000 steps.
All the methods are implemented with PyTorch~\cite{Paszke2019Pytorch} and experimented on a single NVIDIA GTX 3090 GPU. Note that the process of model inference can be further accelerated via better implementations (e.g., using ONNX and TensorRT). In Table~\ref{tab:time_memo} we also report the training time of black-box tuning using ONNX Runtime. Detailed calculation of the amount of data to be uploaded/downloaded can be found in Appendix~\ref{append:data_size}.

\subsection{Ablation Study}
\label{sec:ablation}

In this section, we conduct ablation experiments on various hyper-parameters. To control experimental variables, we explore the effect of each hyper-parameter while keeping the other hyper-parameters as default as listed in Table~\ref{tab:hyperparam}. To stablize the experimental results and reduce the variance over different runs, we conduct ablation experiments in 64-shot setting. Each run is performed on the same data split with different random seeds. Experimental results of ablations on loss functions $\mathcal{L}$, subspace dimensionality $d$, and prompt length $L$ are demonstrated in Figure~\ref{fig:ablation}. Additional ablation studies on the effect of the random projection $\mathbf{A}$, the effect of the population size $\lambda$, and the ablations in the 16-shot setting are in Appendix~\ref{append:add_exp}.

For each ablation, we show results under different budget, which is measured by the number of PTM inference API calls. In each API call, one can provide a continuous prompt $\mathbf{p}$ and query the results of the PTM forward computation on a batch of training data. In our few-shot setting, we can put all the training data into one batch, and therefore the objective function to be optimized is deterministic instead of stochastic.

\paragraph{CMA-ES vs. Adam.} We compare our used derivative-free optimizer, CMA-ES, with a competitive first-order optimizer, Adam~\cite{kingma2015adam}. For fair comparison, we update the continuous prompt using Adam with the gradients over the entire training data (i.e., batch size equals to $|\mathcal{D}_{\text{train}}|$). We use learning rate of 1e-3 for Adam optimizer. As shown in the top row of Figure~\ref{fig:ablation}, Adam optimizer achieves faster convergence on both SST-2 and AG's News due to the gradients it used. On the development sets, Adam performs slight worse than CMA-ES with cross entropy on SST-2 but better on AG's News. But as demonstrated in Table~\ref{tab:main_results}, using Adam optimizer performs worse than CMA-ES on the average performance across seven task test sets.

\paragraph{Loss Functions.} We consider three loss functions: cross entropy, hinge loss, and negative accuracy. As depicted in the top row of Figure~\ref{fig:ablation}, cross entropy and hinge loss significantly outperform the negative accuracy. In the few-shot setting, the accuracy as a reward can be sparse, and cannot provide informative directions for optimization. On SST-2 and AG's News, we obtain that cross entropy performs slightly better than hinge loss.

\paragraph{Subspace Dimensionality.} The subspace of dimensionality $d$ is the space where the optimization actually performs. According to the intrinsic dimensionality found in \citet{Aghajanyan2021Intrinsic}, we explore the subspace dimensionality of \{100, 200, 500, 1000\} within the budget of \{2k, 4k, 6k, 8k\}. Accordingly, we set population size $\lambda=4+3\log(d)$. As shown in the middle row of Figure~\ref{fig:ablation}, the best subspace dimensionality can be different on different tasks ($d=200$ performs the best on SST-2 development set and $d=500$ performs the best on AG's News development set), which is related to the observation that intrinsic dimensionality varies across different tasks~\cite{Aghajanyan2021Intrinsic}. In general, a small subspace (e.g., $d=100$) is hard to cover a good solution, while a large subspace (e.g., $d=1000$) may lead to poor generalization.


\paragraph{Prompt Length.} Prompt length $L$ determines the dimensionality of the original parameter space (in our case $D=L\times 1024$). We evaluate black-box tuning under each budget in \{2k, 4k, 6k, 8k\} while varying the prompt length in \{10, 20, 50, 100\}. As shown in the bottom row of Figure~\ref{fig:ablation}, shorter prompt confers faster convergence on the training sets but does not yield better generalization on the development sets. $L=50$ achieves the best accuracy on both SST-2 and AG's News development sets.



\section{Discussion and Future Work}
In this section we discuss our proposed method in the context of (1) derivative-free optimization and (2) prompt-based learning, respectively. By drawing comparisons with these two lines of research, we highlight some directions that could improve this work in future.

\paragraph{Comparison with Previous Derivative-Free Approaches.}
Our proposed method lies in the same framework of previous work that solves high-dimensional derivative-free optimization problems via random embedding~\cite{Wang2016BORE}. In contrast, we set the random embedding $\mathbf{A}$ by sampling from a uniform distribution instead of normal distributions, and use the CMA-ES to perform optimization in the generated subspace. In previous work, the target black-box functions are usually synthetic functions where only a few dimensions can affect the function values, and therefore most of the dimensions are strictly non-effective. In our real-world scenario, the intrinsic dimension can be approximate. In the context of PTMs, a more appropriate substitution for the term \textit{intrinsic dimensionality} can be \textit{$\epsilon$-effective dimensionality}~\cite{Qian2016DFOHigh}. Considering the relaxation to the intrinsic dimensionality of PTMs, more suitable approaches such as sequential random embedding~\cite{Qian2016DFOHigh} and other more advanced methods of constructing the random projection matrix~\cite{LethamCRB20} should be explored in future work. 
Besides, the subspace generated by random projection can be sub-optimal. As demonstrated in \citet{Qin2021Exploring}, training the projection $\mathbf{A}$ with multi-task supervision can result in better and smaller subspace. Besides, larger PTMs generally have lower intrinsic dimensionalities~\cite{Aghajanyan2021Intrinsic}, as a result, we can use smaller subspace and more efficient DFO algorithms such as Bayesian optimization on larger PTMs. 

\paragraph{Comparison with Previous Prompt-Based Learning Approaches.}
From the perspective of prompt-based learning, our method is similar to prompt-tuning~\cite{Lester2021Prompt}, where only the continuous prompt prepended to the input text is tuned, so our method also retains the benefits of efficient serving and mixed-task inference. In addition to the continuous prompt, we also insert some hard prompt tokens (e.g., "\textit{It was} \texttt{[MASK]}") in the input text, which has been demonstrated to be effective in previous work~\cite{Gu2021PPT} in the name of hybrid prompt tuning. Different from previous prompt-based learning approaches, our prompt tuning does not require backpropagation and gradient descent. Considering our used templates and label words are hand-crafted without trial-and-error, the performance reported in this paper is just a lower bound. More advanced techniques such as prompt engineering~\cite{Gao20Making}, label words engineering~\cite{Schick20Auto,Shin20Autoprompt,Hu21Knowledgeable}, prompt pre-training~\cite{Gu2021PPT}, and prompt ensembling~\cite{Lester2021Prompt} are orthogonal to this work and therefore can further improve the performance. For simplicity, we do not integrate these methods and leave for future work.

\section*{Acknowledgements}
The authors would like to thank Yang Yu for the valuable suggestions of the methods and presentation of the paper, and the anonymous reviewers for their constructive comments. 
This work was supported by the National Key Research and Development Program of China (No. 2020AAA0108702), the National Natural Science Foundation of China (No. 62022027), the major key project of PCL (No. PCL2021A12), and the Natural Science Foundation of Shanghai (No. 21ZR1420300).


\bibliography{example_paper}

\begin{thebibliography}{61}
\providecommand{\natexlab}[1]{#1}
\providecommand{\url}[1]{\texttt{#1}}
\expandafter\ifx\csname urlstyle\endcsname\relax
  \providecommand{\doi}[1]{doi: #1}\else
  \providecommand{\doi}{doi: \begingroup \urlstyle{rm}\Url}\fi

\bibitem[Aghajanyan et~al.(2021)Aghajanyan, Gupta, and
  Zettlemoyer]{Aghajanyan2021Intrinsic}
Aghajanyan, A., Gupta, S., and Zettlemoyer, L.
\newblock Intrinsic dimensionality explains the effectiveness of language model
  fine-tuning.
\newblock In \emph{Proceedings of the 59th Annual Meeting of the Association
  for Computational Linguistics and the 11th International Joint Conference on
  Natural Language Processing, {ACL/IJCNLP} 2021, (Volume 1: Long Papers),
  Virtual Event, August 1-6, 2021}, pp.\  7319--7328, 2021.

\bibitem[Bowman et~al.(2015)Bowman, Angeli, Potts, and Manning]{Bowman2015SNLI}
Bowman, S.~R., Angeli, G., Potts, C., and Manning, C.~D.
\newblock A large annotated corpus for learning natural language inference.
\newblock In \emph{Proceedings of the 2015 Conference on Empirical Methods in
  Natural Language Processing, {EMNLP} 2015, Lisbon, Portugal, September 17-21,
  2015}, pp.\  632--642, 2015.

\bibitem[Brown et~al.(2020)Brown, Mann, Ryder, Subbiah, Kaplan, Dhariwal,
  Neelakantan, Shyam, Sastry, Askell, Agarwal, Herbert{-}Voss, Krueger,
  Henighan, Child, Ramesh, Ziegler, Wu, Winter, Hesse, Chen, Sigler, Litwin,
  Gray, Chess, Clark, Berner, McCandlish, Radford, Sutskever, and
  Amodei]{Brown2020GPT3}
Brown, T.~B., Mann, B., Ryder, N., Subbiah, M., Kaplan, J., Dhariwal, P.,
  Neelakantan, A., Shyam, P., Sastry, G., Askell, A., Agarwal, S.,
  Herbert{-}Voss, A., Krueger, G., Henighan, T., Child, R., Ramesh, A.,
  Ziegler, D.~M., Wu, J., Winter, C., Hesse, C., Chen, M., Sigler, E., Litwin,
  M., Gray, S., Chess, B., Clark, J., Berner, C., McCandlish, S., Radford, A.,
  Sutskever, I., and Amodei, D.
\newblock Language models are few-shot learners.
\newblock In \emph{Advances in Neural Information Processing Systems 33: Annual
  Conference on Neural Information Processing Systems 2020, NeurIPS 2020,
  December 6-12, 2020, virtual}, 2020.

\bibitem[Conn et~al.(2009)Conn, Scheinberg, and Vicente]{DFO-book}
Conn, A.~R., Scheinberg, K., and Vicente, L.~N.
\newblock \emph{Introduction to Derivative-Free Optimization}.
\newblock {SIAM}, Philadelphia, PA, 2009.

\bibitem[Devlin et~al.(2019)Devlin, Chang, Lee, and Toutanova]{Devlin2019BERT}
Devlin, J., Chang, M., Lee, K., and Toutanova, K.
\newblock {BERT:} pre-training of deep bidirectional transformers for language
  understanding.
\newblock In \emph{Proceedings of the 2019 Conference of the North American
  Chapter of the Association for Computational Linguistics: Human Language
  Technologies, {NAACL-HLT} 2019, Minneapolis, MN, USA, June 2-7, 2019, Volume
  1 (Long and Short Papers)}, pp.\  4171--4186, 2019.

\bibitem[Dolan \& Brockett(2005)Dolan and Brockett]{Dolan2005MRPC}
Dolan, W.~B. and Brockett, C.
\newblock Automatically constructing a corpus of sentential paraphrases.
\newblock In \emph{Proceedings of the Third International Workshop on
  Paraphrasing, IWP@IJCNLP 2005, Jeju Island, Korea, October 2005, 2005}, 2005.

\bibitem[Fedus et~al.(2021)Fedus, Zoph, and Shazeer]{Fedus2021Switch}
Fedus, W., Zoph, B., and Shazeer, N.
\newblock Switch transformers: Scaling to trillion parameter models with simple
  and efficient sparsity.
\newblock \emph{arXiv:2101.03961}, 2021.

\bibitem[Gao et~al.(2021)Gao, Fisch, and Chen]{Gao20Making}
Gao, T., Fisch, A., and Chen, D.
\newblock Making pre-trained language models better few-shot learners.
\newblock In \emph{Proceedings of the 59th Annual Meeting of the Association
  for Computational Linguistics and the 11th International Joint Conference on
  Natural Language Processing, {ACL/IJCNLP} 2021, (Volume 1: Long Papers),
  Virtual Event, August 1-6, 2021}, pp.\  3816--3830, 2021.

\bibitem[Gu et~al.(2021)Gu, Han, Liu, and Huang]{Gu2021PPT}
Gu, Y., Han, X., Liu, Z., and Huang, M.
\newblock {PPT:} pre-trained prompt tuning for few-shot learning.
\newblock \emph{arXiv:2109.04332}, 2021.

\bibitem[Hambardzumyan et~al.(2021)Hambardzumyan, Khachatrian, and
  May]{Hambardzumyan20WARP}
Hambardzumyan, K., Khachatrian, H., and May, J.
\newblock {WARP:} word-level adversarial reprogramming.
\newblock In \emph{Proceedings of the 59th Annual Meeting of the Association
  for Computational Linguistics and the 11th International Joint Conference on
  Natural Language Processing, {ACL/IJCNLP} 2021, (Volume 1: Long Papers),
  Virtual Event, August 1-6, 2021}, pp.\  4921--4933, 2021.

\bibitem[Hansen(2016)]{Hansen16a2016CMATutorial}
Hansen, N.
\newblock The {CMA} evolution strategy: {A} tutorial.
\newblock \emph{arXiv:1604.00772}, 2016.

\bibitem[Hansen \& Ostermeier(2001)Hansen and Ostermeier]{Hansen2001CMA}
Hansen, N. and Ostermeier, A.
\newblock Completely derandomized self-adaptation in evolution strategies.
\newblock \emph{Evol. Comput.}, 9\penalty0 (2):\penalty0 159--195, 2001.

\bibitem[Hansen et~al.(2003)Hansen, M{\"{u}}ller, and
  Koumoutsakos]{Hansen2003Reducing}
Hansen, N., M{\"{u}}ller, S.~D., and Koumoutsakos, P.
\newblock Reducing the time complexity of the derandomized evolution strategy
  with covariance matrix adaptation {(CMA-ES)}.
\newblock \emph{Evol. Comput.}, 11\penalty0 (1):\penalty0 1--18, 2003.

\bibitem[He et~al.(2021)He, Zhou, Ma, Berg{-}Kirkpatrick, and
  Neubig]{He2021Unified}
He, J., Zhou, C., Ma, X., Berg{-}Kirkpatrick, T., and Neubig, G.
\newblock Towards a unified view of parameter-efficient transfer learning.
\newblock \emph{arXiv:2110.04366}, 2021.

\bibitem[He et~al.(2015)He, Zhang, Ren, and Sun]{He2015Delving}
He, K., Zhang, X., Ren, S., and Sun, J.
\newblock Delving deep into rectifiers: Surpassing human-level performance on
  imagenet classification.
\newblock In \emph{2015 {IEEE} International Conference on Computer Vision,
  {ICCV} 2015, Santiago, Chile, December 7-13, 2015}, pp.\  1026--1034, 2015.

\bibitem[Hochreiter \& Schmidhuber(1997)Hochreiter and
  Schmidhuber]{Hochreiter1997LSTM}
Hochreiter, S. and Schmidhuber, J.
\newblock Long short-term memory.
\newblock \emph{Neural Comput.}, 9\penalty0 (8):\penalty0 1735--1780, 1997.

\bibitem[Houlsby et~al.(2019)Houlsby, Giurgiu, Jastrzebski, Morrone,
  de~Laroussilhe, Gesmundo, Attariyan, and Gelly]{Houlsby2019Adapter}
Houlsby, N., Giurgiu, A., Jastrzebski, S., Morrone, B., de~Laroussilhe, Q.,
  Gesmundo, A., Attariyan, M., and Gelly, S.
\newblock Parameter-efficient transfer learning for {NLP}.
\newblock In \emph{Proceedings of the 36th International Conference on Machine
  Learning, {ICML} 2019, 9-15 June 2019, Long Beach, California, {USA}},
  volume~97 of \emph{Proceedings of Machine Learning Research}, pp.\
  2790--2799, 2019.

\bibitem[Hu et~al.(2021{\natexlab{a}})Hu, Shen, Wallis, Allen{-}Zhu, Li, Wang,
  and Chen]{Hu2021LoRA}
Hu, E.~J., Shen, Y., Wallis, P., Allen{-}Zhu, Z., Li, Y., Wang, S., and Chen,
  W.
\newblock Lora: Low-rank adaptation of large language models.
\newblock \emph{arXiv:2106.09685}, 2021{\natexlab{a}}.

\bibitem[Hu et~al.(2021{\natexlab{b}})Hu, Ding, Wang, Liu, Li, and
  Sun]{Hu21Knowledgeable}
Hu, S., Ding, N., Wang, H., Liu, Z., Li, J., and Sun, M.
\newblock Knowledgeable prompt-tuning: Incorporating knowledge into prompt
  verbalizer for text classification.
\newblock \emph{arXiv:2108.02035}, 2021{\natexlab{b}}.

\bibitem[Hu et~al.(2017)Hu, Qian, and Yu]{hu2017sequential-aaai17}
Hu, Y.-Q., Qian, H., and Yu, Y.
\newblock Sequential classification-based optimization for direct policy
  search.
\newblock In \emph{Proceedings of the 31st {AAAI} Conference on Artificial
  Intelligence}, pp.\  2029--2035, San Francisco, CA, 2017.

\bibitem[Kingma \& Ba(2015)Kingma and Ba]{kingma2015adam}
Kingma, D.~P. and Ba, J.
\newblock Adam: {A} method for stochastic optimization.
\newblock In \emph{3rd International Conference on Learning Representations,
  {ICLR} 2015, San Diego, CA, USA, May 7-9, 2015, Conference Track
  Proceedings}, 2015.

\bibitem[Kolda et~al.(2003)Kolda, Lewis, and Torczon]{DFO-siamrev}
Kolda, T.~G., Lewis, R.~M., and Torczon, V.
\newblock Optimization by direct search: {N}ew perspectives on some classical
  and modern methods.
\newblock \emph{{SIAM} Review}, 45\penalty0 (3):\penalty0 385--482, 2003.

\bibitem[Lester et~al.(2021)Lester, Al{-}Rfou, and Constant]{Lester2021Prompt}
Lester, B., Al{-}Rfou, R., and Constant, N.
\newblock The power of scale for parameter-efficient prompt tuning.
\newblock In \emph{Proceedings of the 2021 Conference on Empirical Methods in
  Natural Language Processing, {EMNLP} 2021, Virtual Event / Punta Cana,
  Dominican Republic, 7-11 November, 2021}, pp.\  3045--3059, 2021.

\bibitem[Letham et~al.(2020)Letham, Calandra, Rai, and Bakshy]{LethamCRB20}
Letham, B., Calandra, R., Rai, A., and Bakshy, E.
\newblock Re-examining linear embeddings for high-dimensional {B}ayesian
  optimization.
\newblock In \emph{Advances in Neural Information Processing Systems 33},
  virtual, 2020.

\bibitem[Lewis et~al.(2020)Lewis, Liu, Goyal, Ghazvininejad, Mohamed, Levy,
  Stoyanov, and Zettlemoyer]{Lewis2020BART}
Lewis, M., Liu, Y., Goyal, N., Ghazvininejad, M., Mohamed, A., Levy, O.,
  Stoyanov, V., and Zettlemoyer, L.
\newblock {BART:} denoising sequence-to-sequence pre-training for natural
  language generation, translation, and comprehension.
\newblock In \emph{Proceedings of the 58th Annual Meeting of the Association
  for Computational Linguistics, {ACL} 2020, Online, July 5-10, 2020}, pp.\
  7871--7880, 2020.

\bibitem[Li et~al.(2018)Li, Farkhoor, Liu, and Yosinski]{Li2018Intrinsic}
Li, C., Farkhoor, H., Liu, R., and Yosinski, J.
\newblock Measuring the intrinsic dimension of objective landscapes.
\newblock In \emph{6th International Conference on Learning Representations,
  {ICLR} 2018, Vancouver, BC, Canada, April 30 - May 3, 2018, Conference Track
  Proceedings}, 2018.

\bibitem[Li \& Liang(2021)Li and Liang]{Li2021Prefix}
Li, X.~L. and Liang, P.
\newblock Prefix-tuning: Optimizing continuous prompts for generation.
\newblock In \emph{Proceedings of the 59th Annual Meeting of the Association
  for Computational Linguistics and the 11th International Joint Conference on
  Natural Language Processing, {ACL/IJCNLP} 2021, (Volume 1: Long Papers),
  Virtual Event, August 1-6, 2021}, pp.\  4582--4597, 2021.

\bibitem[Liu et~al.(2021{\natexlab{a}})Liu, Ji, Fu, Du, Yang, and
  Tang]{Liu2021PTuningv2}
Liu, X., Ji, K., Fu, Y., Du, Z., Yang, Z., and Tang, J.
\newblock P-tuning v2: Prompt tuning can be comparable to fine-tuning
  universally across scales and tasks.
\newblock \emph{arXiv:2110.07602}, 2021{\natexlab{a}}.

\bibitem[Liu et~al.(2021{\natexlab{b}})Liu, Zheng, Du, Ding, Qian, Yang, and
  Tang]{Liu2021PTuning}
Liu, X., Zheng, Y., Du, Z., Ding, M., Qian, Y., Yang, Z., and Tang, J.
\newblock {GPT} understands, too.
\newblock \emph{arXiv:2103.10385}, 2021{\natexlab{b}}.

\bibitem[Liu et~al.(2019)Liu, Ott, Goyal, Du, Joshi, Chen, Levy, Lewis,
  Zettlemoyer, and Stoyanov]{Liu2019roberta}
Liu, Y., Ott, M., Goyal, N., Du, J., Joshi, M., Chen, D., Levy, O., Lewis, M.,
  Zettlemoyer, L., and Stoyanov, V.
\newblock Roberta: {A} robustly optimized {BERT} pretraining approach.
\newblock \emph{arXiv:1907.11692}, 2019.

\bibitem[Paszke et~al.(2019)Paszke, Gross, Massa, Lerer, Bradbury, Chanan,
  Killeen, Lin, Gimelshein, Antiga, Desmaison, K{\"{o}}pf, Yang, DeVito,
  Raison, Tejani, Chilamkurthy, Steiner, Fang, Bai, and
  Chintala]{Paszke2019Pytorch}
Paszke, A., Gross, S., Massa, F., Lerer, A., Bradbury, J., Chanan, G., Killeen,
  T., Lin, Z., Gimelshein, N., Antiga, L., Desmaison, A., K{\"{o}}pf, A., Yang,
  E., DeVito, Z., Raison, M., Tejani, A., Chilamkurthy, S., Steiner, B., Fang,
  L., Bai, J., and Chintala, S.
\newblock Pytorch: An imperative style, high-performance deep learning library.
\newblock In \emph{Advances in Neural Information Processing Systems 32: Annual
  Conference on Neural Information Processing Systems 2019, NeurIPS 2019,
  December 8-14, 2019, Vancouver, BC, Canada}, pp.\  8024--8035, 2019.

\bibitem[Perez et~al.(2021)Perez, Kiela, and Cho]{Perez2021TrueFewShot}
Perez, E., Kiela, D., and Cho, K.
\newblock True few-shot learning with language models.
\newblock In \emph{Advances in Neural Information Processing Systems 34: Annual
  Conference on Neural Information Processing Systems 2021, NeurIPS 2021,
  December 6-14, 2021, virtual}, pp.\  11054--11070, 2021.

\bibitem[Peters et~al.(2019)Peters, Ruder, and Smith]{Peters2019ToTune}
Peters, M.~E., Ruder, S., and Smith, N.~A.
\newblock To tune or not to tune? adapting pretrained representations to
  diverse tasks.
\newblock In \emph{Proceedings of the 4th Workshop on Representation Learning
  for NLP, RepL4NLP@ACL 2019, Florence, Italy, August 2, 2019}, pp.\  7--14,
  2019.

\bibitem[Qian et~al.(2016)Qian, Hu, and Yu]{Qian2016DFOHigh}
Qian, H., Hu, Y., and Yu, Y.
\newblock Derivative-free optimization of high-dimensional non-convex functions
  by sequential random embeddings.
\newblock In \emph{Proceedings of the Twenty-Fifth International Joint
  Conference on Artificial Intelligence, {IJCAI} 2016, New York, NY, USA, 9-15
  July 2016}, pp.\  1946--1952, 2016.

\bibitem[Qin \& Eisner(2021)Qin and Eisner]{Qin21Learning}
Qin, G. and Eisner, J.
\newblock Learning how to ask: Querying lms with mixtures of soft prompts.
\newblock In \emph{Proceedings of the 2021 Conference of the North American
  Chapter of the Association for Computational Linguistics: Human Language
  Technologies, {NAACL-HLT} 2021, Online, June 6-11, 2021}, pp.\  5203--5212,
  2021.

\bibitem[Qin et~al.(2021)Qin, Wang, Su, Lin, Ding, Liu, Li, Hou, Li, Sun, and
  Zhou]{Qin2021Exploring}
Qin, Y., Wang, X., Su, Y., Lin, Y., Ding, N., Liu, Z., Li, J., Hou, L., Li, P.,
  Sun, M., and Zhou, J.
\newblock Exploring low-dimensional intrinsic task subspace via prompt tuning.
\newblock \emph{arXiv:2110.07867}, 2021.

\bibitem[Qiu et~al.(2020)Qiu, Sun, Xu, Shao, Dai, and Huang]{Qiu2020survey}
Qiu, X., Sun, T., Xu, Y., Shao, Y., Dai, N., and Huang, X.
\newblock Pre-trained models for natural language processing: {A} survey.
\newblock \emph{SCIENCE CHINA Technological Sciences}, 2020.

\bibitem[Raffel et~al.(2020)Raffel, Shazeer, Roberts, Lee, Narang, Matena,
  Zhou, Li, and Liu]{Raffel2020T5}
Raffel, C., Shazeer, N., Roberts, A., Lee, K., Narang, S., Matena, M., Zhou,
  Y., Li, W., and Liu, P.~J.
\newblock Exploring the limits of transfer learning with a unified text-to-text
  transformer.
\newblock \emph{J. Mach. Learn. Res.}, 21:\penalty0 140:1--140:67, 2020.

\bibitem[Rios \& Sahinidis(2013)Rios and Sahinidis]{DFO-review}
Rios, L.~M. and Sahinidis, N.~V.
\newblock Derivative-free optimization: {A} review of algorithms and comparison
  of software implementations.
\newblock \emph{Journal of Global Optimization}, 56\penalty0 (3):\penalty0
  1247--1293, 2013.

\bibitem[Salimans et~al.(2017)Salimans, Ho, Chen, Sidor, and
  Sutskever]{salimans2017evolution}
Salimans, T., Ho, J., Chen, X., Sidor, S., and Sutskever, I.
\newblock Evolution strategies as a scalable alternative to reinforcement
  learning.
\newblock \emph{arXiv:1703.03864}, 2017.

\bibitem[Schick \& Sch{\"{u}}tze(2021{\natexlab{a}})Schick and
  Sch{\"{u}}tze]{Schick21PET}
Schick, T. and Sch{\"{u}}tze, H.
\newblock Exploiting cloze-questions for few-shot text classification and
  natural language inference.
\newblock In \emph{Proceedings of the 16th Conference of the European Chapter
  of the Association for Computational Linguistics: Main Volume, {EACL} 2021,
  Online, April 19 - 23, 2021}, pp.\  255--269, 2021{\natexlab{a}}.

\bibitem[Schick \& Sch{\"{u}}tze(2021{\natexlab{b}})Schick and
  Sch{\"{u}}tze]{Schick21Size}
Schick, T. and Sch{\"{u}}tze, H.
\newblock It's not just size that matters: Small language models are also
  few-shot learners.
\newblock In \emph{Proceedings of the 2021 Conference of the North American
  Chapter of the Association for Computational Linguistics: Human Language
  Technologies, {NAACL-HLT} 2021, Online, June 6-11, 2021}, pp.\  2339--2352,
  2021{\natexlab{b}}.

\bibitem[Schick et~al.(2020)Schick, Schmid, and Sch{\"{u}}tze]{Schick20Auto}
Schick, T., Schmid, H., and Sch{\"{u}}tze, H.
\newblock Automatically identifying words that can serve as labels for few-shot
  text classification.
\newblock In \emph{Proceedings of the 28th International Conference on
  Computational Linguistics, {COLING} 2020, Barcelona, Spain (Online), December
  8-13, 2020}, pp.\  5569--5578, 2020.

\bibitem[Shahriari et~al.(2016)Shahriari, Swersky, Wang, Adams, and
  de~Freitas]{reviewBO16}
Shahriari, B., Swersky, K., Wang, Z., Adams, R.~P., and de~Freitas, N.
\newblock Taking the human out of the loop: {A} review of {B}ayesian
  optimization.
\newblock \emph{Proceedings of the {IEEE}}, 104\penalty0 (1):\penalty0
  148--175, 2016.

\bibitem[Shin et~al.(2020)Shin, Razeghi, IV, Wallace, and
  Singh]{Shin20Autoprompt}
Shin, T., Razeghi, Y., IV, R. L.~L., Wallace, E., and Singh, S.
\newblock Autoprompt: Eliciting knowledge from language models with
  automatically generated prompts.
\newblock In \emph{Proceedings of the 2020 Conference on Empirical Methods in
  Natural Language Processing, {EMNLP} 2020, Online, November 16-20, 2020},
  pp.\  4222--4235, 2020.

\bibitem[Snoek et~al.(2012)Snoek, Larochelle, and Adams]{SnoekLA-nips12}
Snoek, J., Larochelle, H., and Adams, R.~P.
\newblock Practical {B}ayesian optimization of machine learning algorithms.
\newblock In \emph{Advances in Neural Information Processing Systems 25}, pp.\
  2960--2968, Lake Tahoe, NV, 2012.

\bibitem[Socher et~al.(2013)Socher, Perelygin, Wu, Chuang, Manning, Ng, and
  Potts]{Socher2013SST}
Socher, R., Perelygin, A., Wu, J., Chuang, J., Manning, C.~D., Ng, A.~Y., and
  Potts, C.
\newblock Recursive deep models for semantic compositionality over a sentiment
  treebank.
\newblock In \emph{Proceedings of the 2013 Conference on Empirical Methods in
  Natural Language Processing, {EMNLP} 2013, 18-21 October 2013, Grand Hyatt
  Seattle, Seattle, Washington, USA, {A} meeting of SIGDAT, a Special Interest
  Group of the {ACL}}, pp.\  1631--1642, 2013.

\bibitem[Sun et~al.(2022)Sun, Liu, Qiu, and Huang]{Sun2021Paradigm}
Sun, T., Liu, X., Qiu, X., and Huang, X.
\newblock Paradigm shift in natural language processing.
\newblock \emph{Machine Intelligence Research}, 2022.

\bibitem[Sun et~al.(2021)Sun, Wang, Feng, Ding, Pang, Shang, Liu, Chen, Zhao,
  Lu, Liu, Wu, Gong, Liang, Shang, Sun, Liu, Ouyang, Yu, Tian, Wu, and
  Wang]{Sun2021ERNIE3}
Sun, Y., Wang, S., Feng, S., Ding, S., Pang, C., Shang, J., Liu, J., Chen, X.,
  Zhao, Y., Lu, Y., Liu, W., Wu, Z., Gong, W., Liang, J., Shang, Z., Sun, P.,
  Liu, W., Ouyang, X., Yu, D., Tian, H., Wu, H., and Wang, H.
\newblock {ERNIE} 3.0: Large-scale knowledge enhanced pre-training for language
  understanding and generation.
\newblock \emph{arXiv:2107.02137}, 2021.

\bibitem[Wang et~al.(2019)Wang, Singh, Michael, Hill, Levy, and
  Bowman]{Wang2019GLUE}
Wang, A., Singh, A., Michael, J., Hill, F., Levy, O., and Bowman, S.~R.
\newblock {GLUE:} {A} multi-task benchmark and analysis platform for natural
  language understanding.
\newblock In \emph{7th International Conference on Learning Representations,
  {ICLR} 2019, New Orleans, LA, USA, May 6-9, 2019}, 2019.

\bibitem[Wang et~al.(2016)Wang, Hutter, Zoghi, Matheson, and
  de~Freitas]{Wang2016BORE}
Wang, Z., Hutter, F., Zoghi, M., Matheson, D., and de~Freitas, N.
\newblock Bayesian optimization in a billion dimensions via random embeddings.
\newblock \emph{J. Artif. Intell. Res.}, 55:\penalty0 361--387, 2016.

\bibitem[Weston \& Watkins(1999)Weston and Watkins]{Weston1999SVMMC}
Weston, J. and Watkins, C.
\newblock Support vector machines for multi-class pattern recognition.
\newblock In \emph{{ESANN} 1999, 7th European Symposium on Artificial Neural
  Networks, Bruges, Belgium, April 21-23, 1999, Proceedings}, pp.\  219--224,
  1999.

\bibitem[Williams et~al.(2018)Williams, Nangia, and Bowman]{Williams2018MNLI}
Williams, A., Nangia, N., and Bowman, S.~R.
\newblock A broad-coverage challenge corpus for sentence understanding through
  inference.
\newblock In \emph{Proceedings of the 2018 Conference of the North American
  Chapter of the Association for Computational Linguistics: Human Language
  Technologies, {NAACL-HLT} 2018, New Orleans, Louisiana, USA, June 1-6, 2018,
  Volume 1 (Long Papers)}, pp.\  1112--1122, 2018.

\bibitem[Wu et~al.(2021)Wu, Zhao, Yu, Zhang, Shen, Liu, Li, Zhu, Luo, Xu, and
  Zhang]{Wu2021Yuan}
Wu, S., Zhao, X., Yu, T., Zhang, R., Shen, C., Liu, H., Li, F., Zhu, H., Luo,
  J., Xu, L., and Zhang, X.
\newblock Yuan 1.0: Large-scale pre-trained language model in zero-shot and
  few-shot learning.
\newblock \emph{arXiv:2110.04725}, 2021.

\bibitem[Zeng et~al.(2021)Zeng, Ren, Su, Wang, Liao, Wang, Jiang, Yang, Wang,
  Zhang, Li, Gong, Yao, Huang, Wang, Yu, Guo, Yu, Zhang, Wang, Tao, Yan, Yi,
  Peng, Jiang, Zhang, Deng, Zhang, Lin, Zhang, Zhang, Guo, Gu, Fan, Wang, Jin,
  Liu, and Tian]{Zeng2021PanGu}
Zeng, W., Ren, X., Su, T., Wang, H., Liao, Y., Wang, Z., Jiang, X., Yang, Z.,
  Wang, K., Zhang, X., Li, C., Gong, Z., Yao, Y., Huang, X., Wang, J., Yu, J.,
  Guo, Q., Yu, Y., Zhang, Y., Wang, J., Tao, H., Yan, D., Yi, Z., Peng, F.,
  Jiang, F., Zhang, H., Deng, L., Zhang, Y., Lin, Z., Zhang, C., Zhang, S.,
  Guo, M., Gu, S., Fan, G., Wang, Y., Jin, X., Liu, Q., and Tian, Y.
\newblock Pangu-{\(\alpha\)}: Large-scale autoregressive pretrained chinese
  language models with auto-parallel computation.
\newblock \emph{arXiv:2104.12369}, 2021.

\bibitem[Zhang et~al.(2021{\natexlab{a}})Zhang, Wu, Katiyar, Weinberger, and
  Artzi]{Zhang2021Revisiting}
Zhang, T., Wu, F., Katiyar, A., Weinberger, K.~Q., and Artzi, Y.
\newblock Revisiting few-sample {BERT} fine-tuning.
\newblock In \emph{9th International Conference on Learning Representations,
  {ICLR} 2021, Virtual Event, Austria, May 3-7, 2021}, 2021{\natexlab{a}}.

\bibitem[Zhang et~al.(2015{\natexlab{a}})Zhang, Zhao, and LeCun]{Zhang2015Char}
Zhang, X., Zhao, J.~J., and LeCun, Y.
\newblock Character-level convolutional networks for text classification.
\newblock In \emph{Advances in Neural Information Processing Systems 28: Annual
  Conference on Neural Information Processing Systems 2015, December 7-12,
  2015, Montreal, Quebec, Canada}, pp.\  649--657, 2015{\natexlab{a}}.

\bibitem[Zhang et~al.(2015{\natexlab{b}})Zhang, Sohn, Villegas, Pan, and
  Lee]{cvpr-ZhangSVPL15}
Zhang, Y., Sohn, K., Villegas, R., Pan, G., and Lee, H.
\newblock Improving object detection with deep convolutional networks via
  {B}ayesian optimization and structured prediction.
\newblock In \emph{{IEEE} Conference on Computer Vision and Pattern
  Recognition}, pp.\  249--258, Boston, MA, 2015{\natexlab{b}}.

\bibitem[Zhang et~al.(2020)Zhang, Han, Zhou, Ke, Gu, Ye, Qin, Su, Ji, Guan, Qi,
  Wang, Zheng, Zeng, Cao, Chen, Li, Sun, Liu, Huang, Han, Tang, Li, Zhu, and
  Sun]{Zhang2020CPM}
Zhang, Z., Han, X., Zhou, H., Ke, P., Gu, Y., Ye, D., Qin, Y., Su, Y., Ji, H.,
  Guan, J., Qi, F., Wang, X., Zheng, Y., Zeng, G., Cao, H., Chen, S., Li, D.,
  Sun, Z., Liu, Z., Huang, M., Han, W., Tang, J., Li, J., Zhu, X., and Sun, M.
\newblock {CPM:} {A} large-scale generative chinese pre-trained language model.
\newblock \emph{arXiv:2012.00413}, 2020.

\bibitem[Zhang et~al.(2021{\natexlab{b}})Zhang, Gu, Han, Chen, Xiao, Sun, Yao,
  Qi, Guan, Ke, Cai, Zeng, Tan, Liu, Huang, Han, Liu, Zhu, and
  Sun]{Zhang2021CPM2}
Zhang, Z., Gu, Y., Han, X., Chen, S., Xiao, C., Sun, Z., Yao, Y., Qi, F., Guan,
  J., Ke, P., Cai, Y., Zeng, G., Tan, Z., Liu, Z., Huang, M., Han, W., Liu, Y.,
  Zhu, X., and Sun, M.
\newblock {CPM-2:} large-scale cost-effective pre-trained language models.
\newblock \emph{arXiv:2106.10715}, 2021{\natexlab{b}}.

\bibitem[Zhong et~al.(2021)Zhong, Friedman, and Chen]{Zhong21OptiPrompt}
Zhong, Z., Friedman, D., and Chen, D.
\newblock Factual probing is {[MASK]:} learning vs. learning to recall.
\newblock In \emph{Proceedings of the 2021 Conference of the North American
  Chapter of the Association for Computational Linguistics: Human Language
  Technologies, {NAACL-HLT} 2021, Online, June 6-11, 2021}, pp.\  5017--5033,
  2021.

\end{thebibliography}
\bibliographystyle{icml2022}
\newpage

\newpage
\appendix
\onecolumn
\section{Additional Experimental Results}
\label{append:add_exp}
\begin{figure}[h]
    \centering
    \subfigure{
    \includegraphics[width=.36\linewidth]{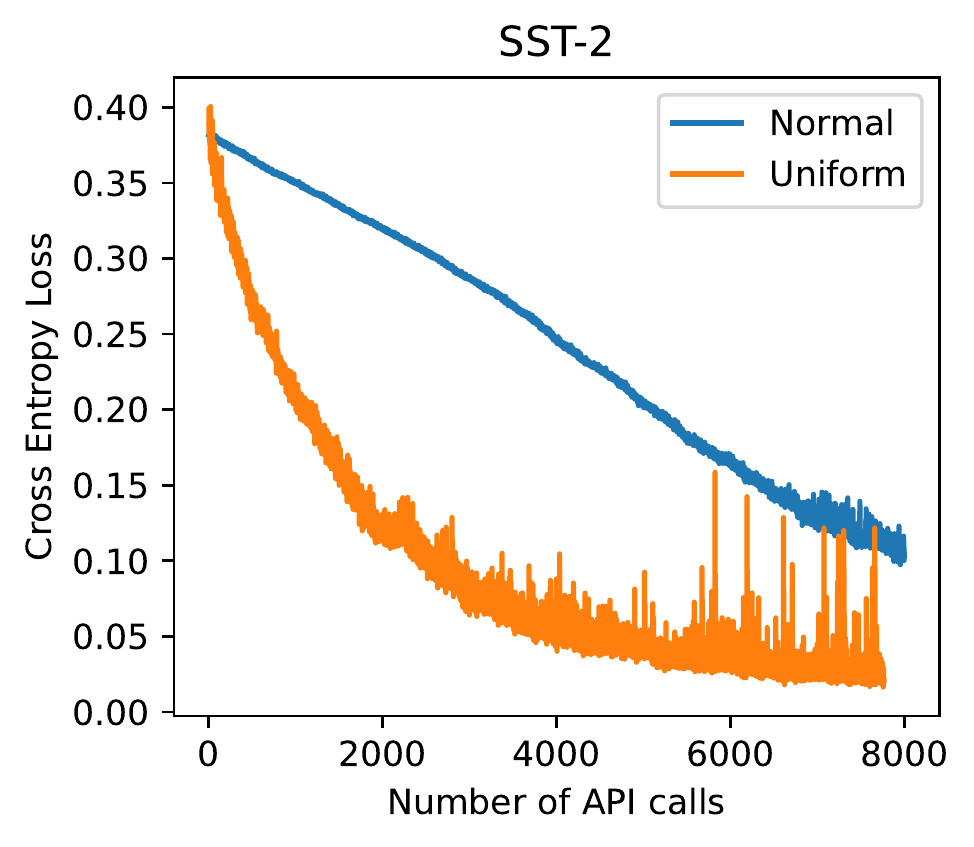}
    }
    \subfigure{
    \includegraphics[width=.36\linewidth]{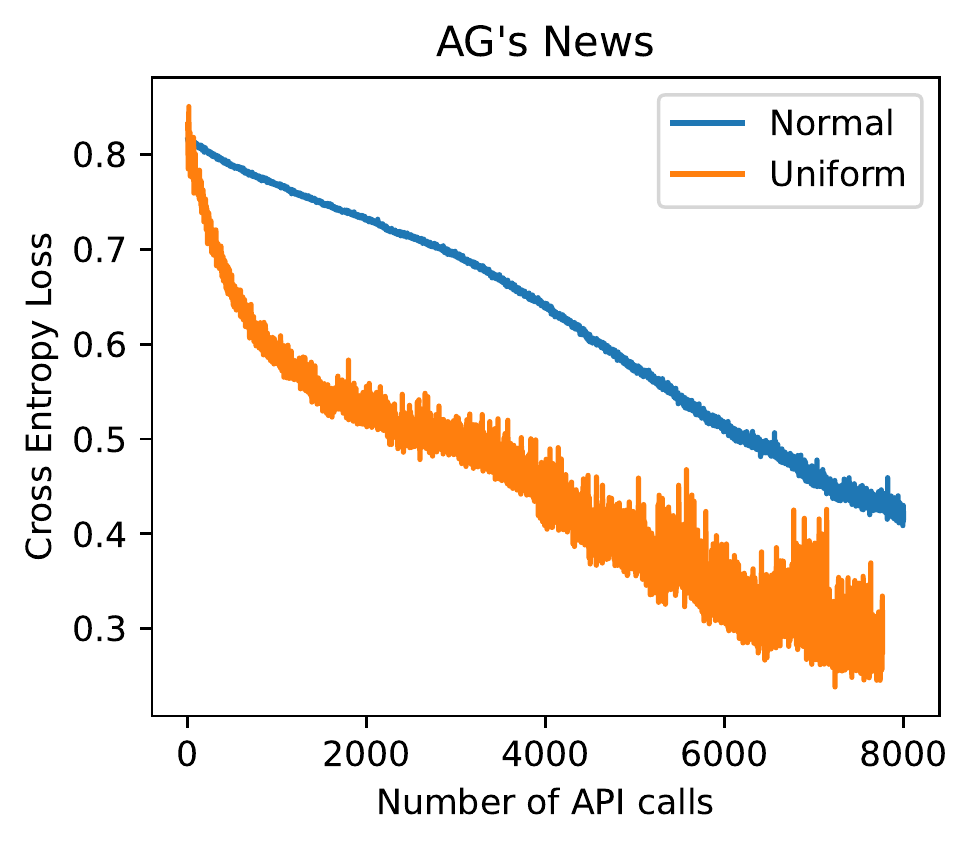}
    }
    \caption{Effect of random projection $\mathbf{A}$.}
    \label{fig:re}
\end{figure}

\paragraph{Random Projection.} 
The random projection matrix $\mathbf{A}\in\mathbb{R}^{D\times d}$ is a key factor that determines whether and how hard it is to find a good solution in the generated subspace. Here we compare two design choices of setting $\mathbf{A}$: The first choice is commonly used in previous high-dimensional derivative-free optimization work~\cite{Wang2016BORE,Qian2016DFOHigh}, that is setting each entry of $\mathbf{A}$ by sampling from a normal distribution. Following \citet{Qian2016DFOHigh}, we use $\mathcal{N}(0, 1/d)$ where $d$ is the subspace dimensionality\footnote{We also tried $\mathcal{N}(0,1)$ as used in \citet{Wang2016BORE}, which does not work in our case. For $\mathcal{N}(0, 1/d)$, we adopt a larger search space $\mathcal{Z}$ instead of $[-5,5]^d$ to get it work.}. The second choice is setting each entry of $\mathbf{A}$ by sampling from a uniform distribution, which is widely used for initializing linear layers in modern neural networks. Here we use the uniform distribution proposed in \citet{He2015Delving}. As shown in Figure~\ref{fig:re}, both random projections can achieve a considerable cross entropy loss on SST-2 and AG's News within reasonable budgets but faster convergence is obtained using uniform distribution.

\begin{figure}[h]
    \centering
    \subfigure{
    \includegraphics[width=.36\linewidth]{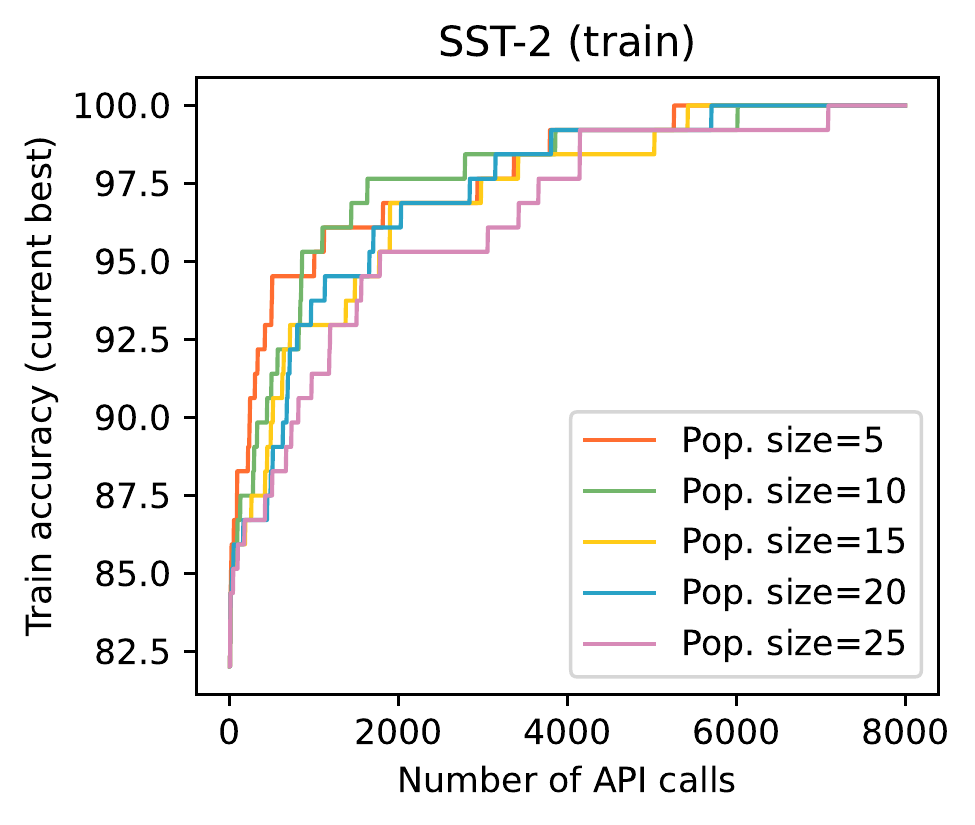}
    }
    \subfigure{
    \includegraphics[width=.36\linewidth]{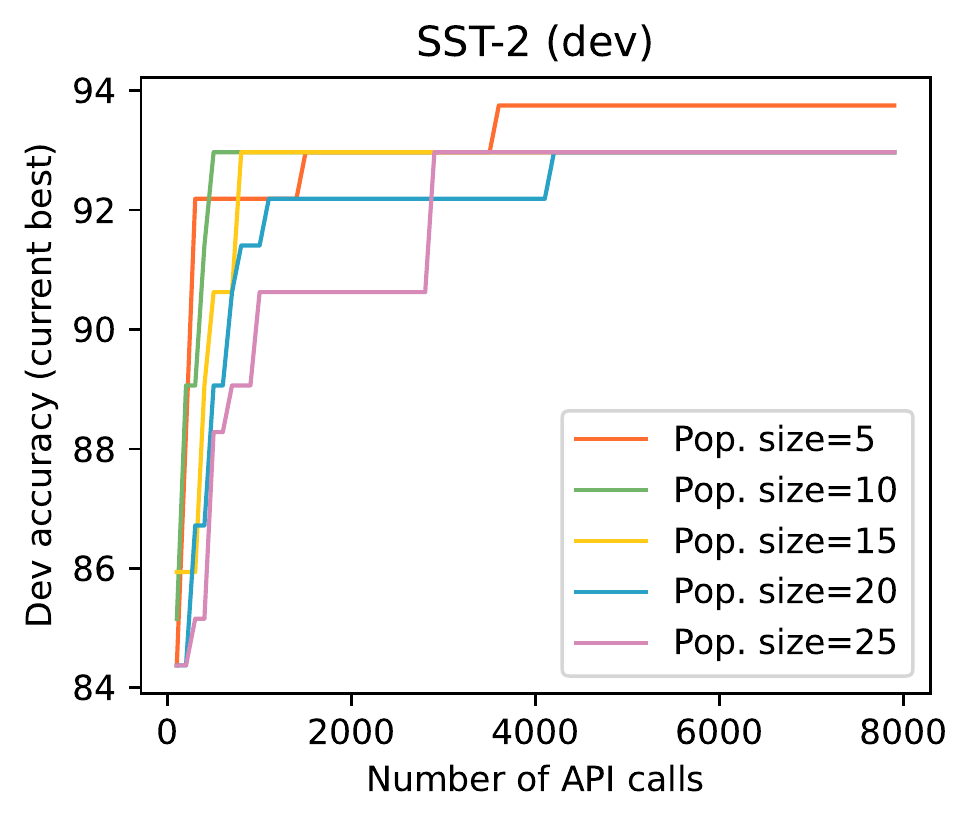}
    }
    \caption{Effect of population size $\lambda$.}
    \label{fig:popsize}
\end{figure}

\paragraph{Population Size.}
In each iteration of the CMA-ES, a population of solutions are sampled from a multivariate normal distribution model. The evaluation of the population is then used to update the parameters of the multivariate normal distribution model. Here we study the effect of the population size on SST-2. In our experiments, we sequentially evaluate each solution in a population, and therefore larger population size will result in more API calls given the same CMA-ES iterations. As shown in Figure~\ref{fig:popsize}, smaller population size confers faster convergence in terms of number of API calls. We also demonstrate the comparison in terms of the CMA-ES iterations, which can be found in the following section.

\begin{figure}[t]
    \centering
    \subfigure{
    \includegraphics[height=.23\linewidth]{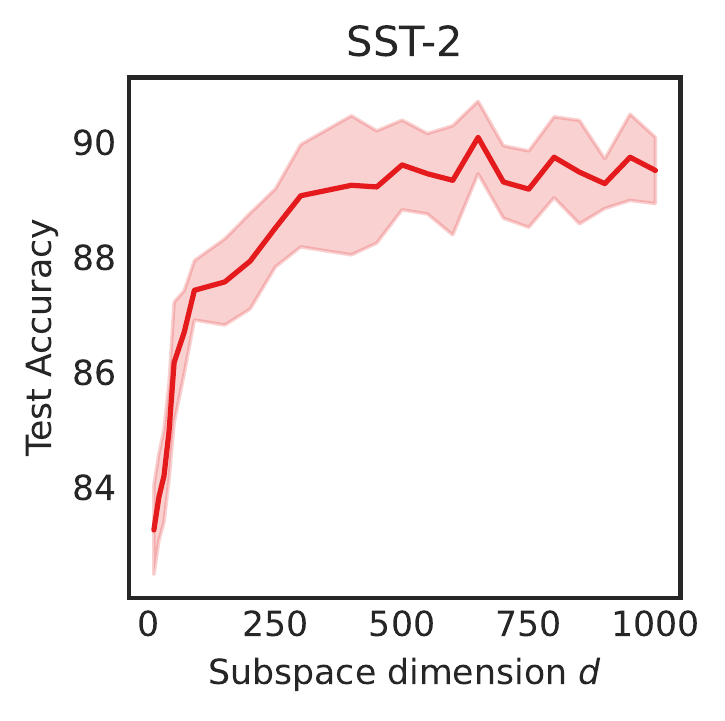}
    }
    \subfigure{
    \includegraphics[height=.23\linewidth]{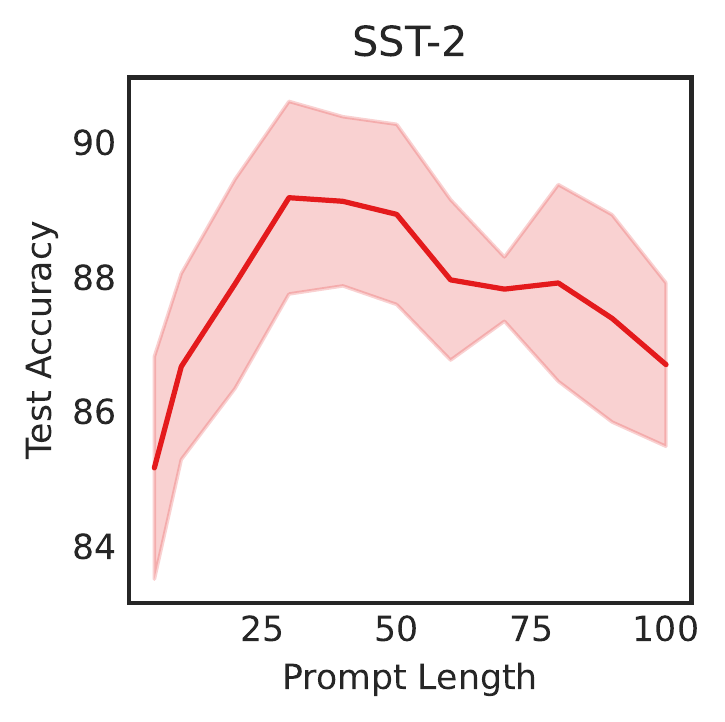}
    }
    \subfigure{
    \includegraphics[height=.23\linewidth]{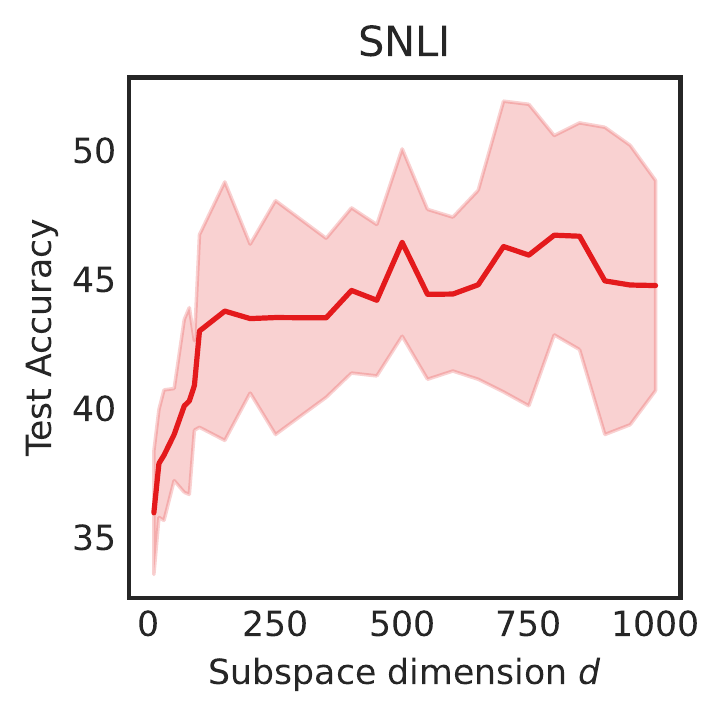}
    }
    \subfigure{
    \includegraphics[height=.23\linewidth]{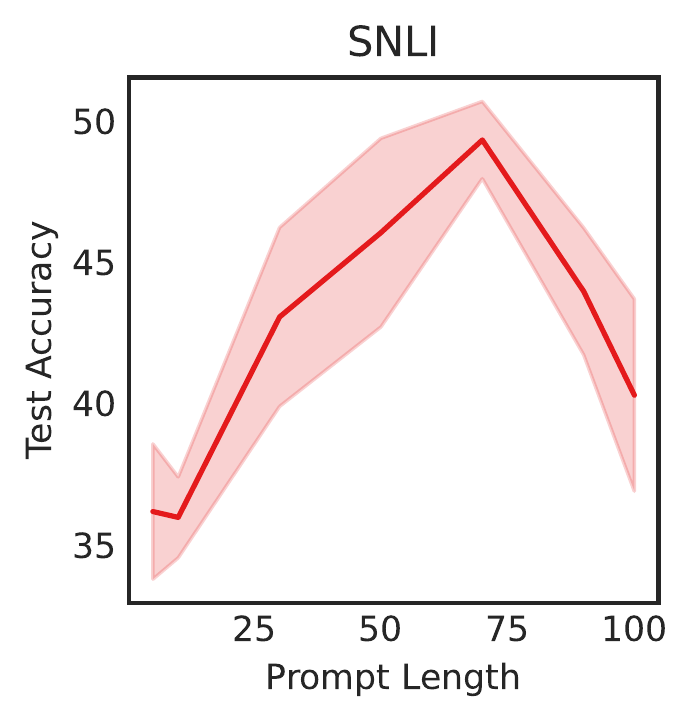}
    }
    \caption{Ablation of subspace dimensionality and prompt length in 16-shot setting.}
    \label{fig:ablation16shot}
\end{figure}

\paragraph{Ablation of Subspace Dimensionality and Prompt Length in 16-shot Setting.}
In \cref{sec:ablation}, we conduct ablation experiments in the 64-shot setting to reduce the variance over different runs. To keep consistent with the experimental setting in Table~\ref{tab:main_results}, we demonstrate in Figure~\ref{fig:ablation16shot} the ablation results on subspace dimensionality and prompt length in the 16-shot setting.

\begin{figure}[t]
    \centering
    \includegraphics[width=\linewidth]{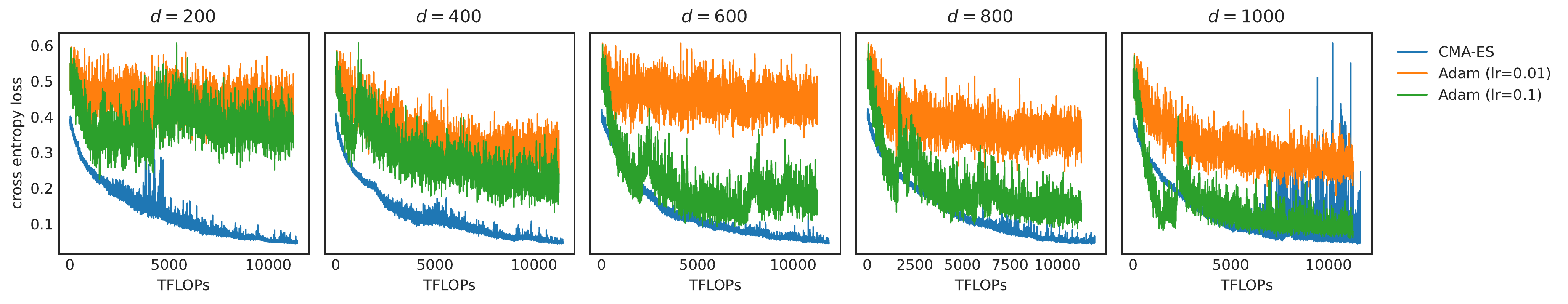}
    \caption{Optimization in low-dimensional subspaces using CMA-ES and Adam.}
    \label{fig:cma_vs_adam}
\end{figure}

\paragraph{CMA-ES vs. Adam in Subspaces.}
In Figure~\ref{fig:ablation}, we compare the convergence of prompt tuning (with Adam optimizer) and black-box tuning (with CMA-ES), where Adam performs optimization in the original prompt space ($\mathcal{P}$) while CMA-ES performs in the generated subsapce ($\mathcal{Z}$). Here we also compare the effectiveness and efficiency of Adam and CMA-ES in subspaces. As shown in Figure~\ref{fig:cma_vs_adam}, CMA-ES is more efficient and stable than Adam in low-dimensional subspaces. When the dimensionality of the subsapce becomes large (e.g., $d=1000$), Adam with a appropriate learning rate can perform on par with CMA-ES. Note that CMA-ES does not require back-propagation, so the computation cost of one iteration for CMA-ES and Adam can be very different. For fair comparison, we convert the number of iterations into FLOPs. The FLOPs of one iteration of Adam is estimated to be three times greater than CMA-ES.

\section{Parallel Evaluation}
\label{append:parallel}
\begin{figure}[h]
    \centering
    \subfigure[]{
    \includegraphics[width=.4\linewidth]{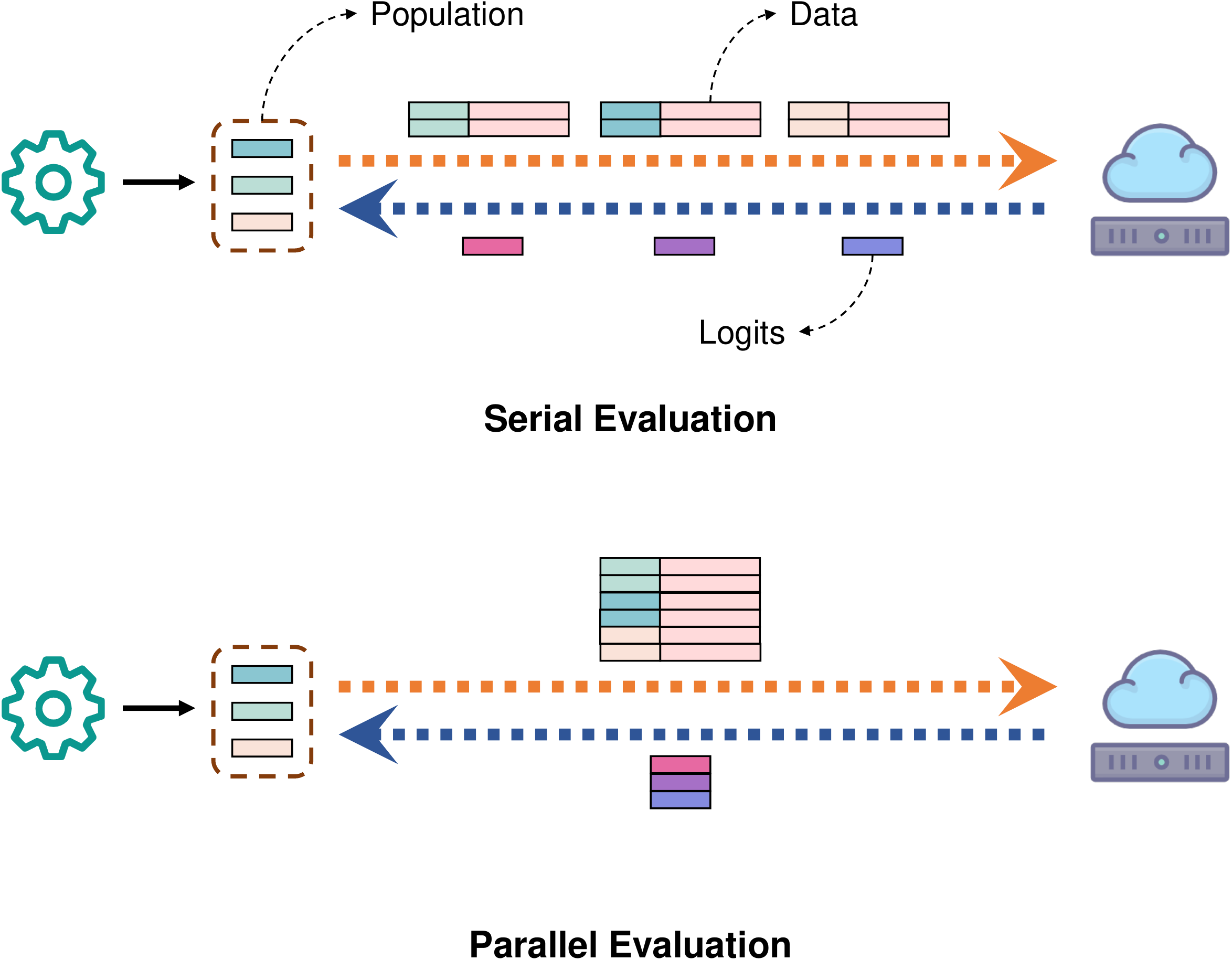}
    }
    \subfigure[]{
    \includegraphics[width=.4\linewidth]{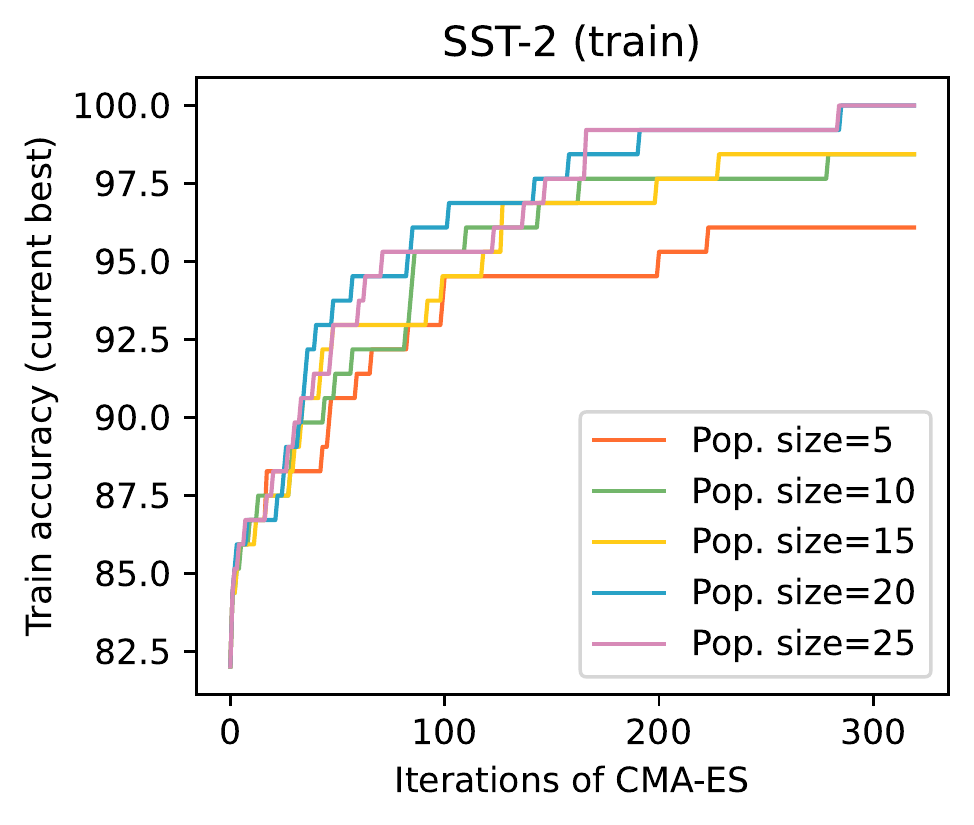}
    }
    \caption{(a) Illustration of the parallel evaluation. (b) Comparison of the convergence rate with different population sizes using parallel evaluation.}
    \label{fig:parallel}
\end{figure}

If the training data is smaller, or the server allows larger batches, a promising way to improve training efficiency is to use parallel evaluation. That is, we can evaluate the entire population in parallel, as depicted in Figure~\ref{fig:parallel}(a). As demonstrated in Figure~\ref{fig:parallel}(b), we can achieve 100\% accuracy on the SST-2 training set with population size of 20 and 25 in 300 iterations (API calls). In case of the batch size per API call is limited, we can also use asynchronous queries to simulate the parallel evaluation.

\section{Estimation of Uploaded/Downloaded Data Size}
\label{append:data_size}
In this section we describe how we estimate the amount of data to be uploaded and downloaded (Table~\ref{tab:time_memo}).

For black-box tuning, there are two kinds of data to be uploaded: (1) training samples, and (2) continuous prompt. A training sample is comprised of two parts: \texttt{input\_ids} and \texttt{attention\_mask}. We can use the unsigned short (representation range: 0$\sim$65535, 2 bytes per value) for \texttt{input\_ids} and use the bool type (1 byte per value) for \texttt{attention\_mask}. For continuous prompt, which contains hundreds of values, we can use the float type (4 bytes per value) for representation. Take SST-2 16-shot split as an example, the \texttt{input\_ids} and \texttt{attention\_mask} are in shape of $32\times 47$, where 32 is the batch size and 47 is the maximum sequence length, so there are $\sim$2.9KB data for \texttt{input\_ids} and $\sim$1.5KB data for \texttt{attention\_mask}. Assume the prompt is 500-dimensional, we need to upload additional $\sim$2KB data for prompt. The data to be downloaded is the output logits of the candidate words, which is a dictionary containing $\mid\mathcal{Y}\mid$ float values. Take SST-2 16-shot split as an example, the size of data to be downloaded is $32\times2\times4\text{bytes}=0.25$KB.

For feature-based methods we use similar estimation methods. The data size for upload is the same for Feature-MLP and Feature-BiLSTM. The data to be downloaded for Feature-MLP is the representation of the \texttt{[CLS]} token while the data to be downloaded for Feature-BiLSTM is the representation of all the tokens. Note that this estimation, without any data compression, is an upper bound of the real scenario. 


\end{document}